\pdfoutput=1
\documentclass[11pt]{article}

\usepackage{EMNLP2022}

\usepackage{times}
\usepackage{latexsym}

\usepackage[T1]{fontenc}
\usepackage[utf8]{inputenc}

\usepackage{microtype}
\usepackage{amsmath}
\usepackage{graphicx}
\usepackage{multirow}
\usepackage{soul}

\usepackage{pifont}

\usepackage{times}
\usepackage{latexsym}
\usepackage{graphicx}
\usepackage{booktabs}
\usepackage{subcaption}
\usepackage[ruled,vlined]{algorithm2e}
\usepackage{multirow}
\usepackage{tabularx}
\usepackage{amsmath}
\usepackage{bbold}
\usepackage{mathtools}
\usepackage{xcolor}
\usepackage{booktabs}
\usepackage{tabularx,ragged2e}
\usepackage{enumitem}
\usepackage{inconsolata}

\title{Model Cascading: Towards Jointly Improving Efficiency and Accuracy of NLP Systems}

\author{Neeraj Varshney and 
  Chitta Baral
  \\
  Arizona State University \\
  \texttt{\{nvarshn2, cbaral\}}@asu.edu
  }
\begin{document}
\maketitle
\begin{abstract}
\textit{Do all instances need inference through the big models for a correct prediction?}\\
Perhaps not; some instances are easy and can be answered correctly by even small capacity models.
This provides opportunities for improving the computational efficiency of systems.
In this work, we present an explorative study on `model cascading', a simple technique that utilizes a collection of models of varying capacities to accurately yet efficiently output predictions.
Through comprehensive experiments in multiple task settings that differ in the number of models available for cascading ($K$ value), we show that cascading improves both the computational efficiency and the prediction accuracy.
For instance, in K=3 setting, cascading saves up to $88.93\%$ computation cost and consistently achieves superior prediction accuracy with an improvement of up to $2.18\%$.
We also study the impact of introducing additional models in the cascade and show that it further increases the efficiency improvements.
Finally, we hope that our work will facilitate development of efficient NLP systems making their widespread adoption in real-world applications possible.
\end{abstract}

\section{Introduction}
Pre-trained language models such as RoBERTa \cite{Liu2019RoBERTaAR}, ELECTRA \cite{clark2020electra}, and T5 \cite{JMLR:v21:20-074} have achieved remarkable performance on numerous natural language processing benchmarks \cite{wang-etal-2018-glue, NEURIPS2019_4496bf24, talmor-etal-2019-commonsenseqa}.
However, these models have a large number of parameters which makes them slow and computationally expensive; for instance, T5-11B requires  ${\sim}87$ $\times$ $10^{11}$ floating point operations (FLOPs) for an inference. 
This limits their widespread adoption in real-world applications that prefer computationally efficient systems in order to achieve low response times.

\begin{figure}[t!]
    \centering
    \includegraphics[width=7.5cm]{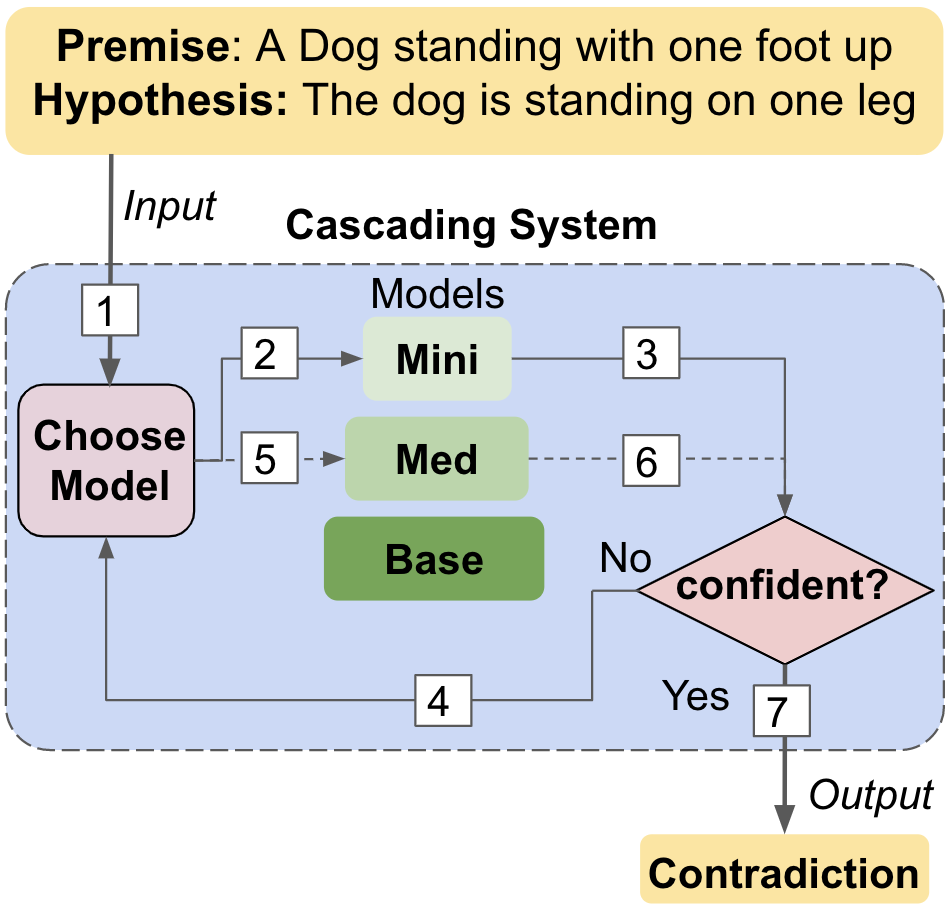}
    \caption{Illustrating a cascading approach with three models (Mini, Med, and Base) arranged in increasing order of capacity. An input is first passed through the smallest model (Mini) which fails to predict with sufficient confidence. Therefore, it is then inferred using a bigger model (Med) that satisfies the confidence constraints and the system outputs its prediction (`contradiction' as dog has four legs). Thus, by avoiding inference through large/expensive models, the system saves computation cost without sacrificing the accuracy.}
    
    \label{fig:teaser}
\end{figure}

The above concern has recently received considerable attention from the NLP community leading to development of several techniques,
such as (1) \textit{network pruning} that progressively removes model weights from a big network
% resulting in a smaller network 
\cite{wang-etal-2020-structured,guo-etal-2021-parameter}, 
% (2) \textit{quantization} that reduces the number of unique values required to represent model weights and activations \cite{shen2020q,zhang-etal-2020-ternarybert},
(2) \textit{early exiting} that allows multiple exit paths in a model \cite{xin-etal-2020-deebert},
(3) \textit{adaptive inference} that adjusts model size by adaptively selecting its width and depth \cite{goyal2020power,kim-cho-2021-length},
(4) \textit{knowledge distillation} that transfers `dark-knowledge' from a large teacher model to a shallow student model \cite{jiao-etal-2020-tinybert,li-etal-2022-dq}, and
(5) \textit{input reduction} that eliminates less contributing tokens from the input text to speed up inference \cite{modarressi-etal-2022-adapler}.
These methods typically require architectural modifications, network manipulation, saliency quantification, or even complex training procedures.
Moreover, computational efficiency in these methods often comes with a compromise on accuracy.
In contrast, \textit{model cascading}, a simple technique that utilizes a collection of models of varying capacities to \textbf{accurately yet efficiently} output predictions has remained underexplored.
% Several smaller models have been developed \cite{turc2019well} but their accuracy is compromised.
% in which a large model is compressed to a lightweight smaller model \cite{DBLP:conf/icml/LiWSLKKG20, 10.1162/tacl_a_00413}, ..., 

In this work, we address the above limitation by first providing mathematical formulation of model cascading and then exploring several approaches to do it.
In its problem setup, a collection of models of different capacities (and hence performances) are provided and the system needs to output its prediction by leveraging one or more models.
On one extreme, the system can use only the smallest model and on the other extreme, it can use all the available models (ensembling).
The former system would be highly efficient but usually poor in performance while the latter system would be fairly accurate but expensive in computation.
Model cascading strives to get the best of both worlds by allowing the system to efficiently utilize the available models while achieving high prediction accuracy.
This is in line with the `Efficiency NLP' \cite{AraseEfficientNP} policy document put up by the ACL community.
% The former scenario would render the system highly efficient but with poor accuracy.
% will usually have poor accuracy.
% its accuracy would usually be poor.
% The latter scenario corresponds to \textit{ensembling} and that system would be quite accurate but computationally very expensive.
% It provides efficiency benefits because 

% Furthermore, some input may not be trivial but can be answered correctly by moderately sized models if not by the smallest model. 
% Thus, computationally intensive inference through large models can be avoided .
% While in the latter scenario, the system 
% would require huge amount of computation. 
% though, using large models for such inputs would also result in correct predictions but would consume significantly higher computation. 
% Model cascading task pertains to striking the balance between accuracy and efficiency trade-offs.

% using the smaller model for a large number of instances and the larger model for the remaining.
% a large number of samples can be answered using the smaller model and only a few samples require inference from the larger model for a correct prediction.
% Intuitively, not all the instances require inference from large models to be answered correctly i.e. some instances are trivial and can be answered correctly by even small capacity models. 
% This allows the system to save computation cost without sacrificing the accuracy.
Consider the case of CommitmentBank \cite{Marneffe2019TheCI} dataset on which BERT-medium model having just 41.7M parameters achieves $75\%$ accuracy and a bigger model BERT-base having 110M parameters achieves $82\%$  accuracy.
From this, it is clear that the performance of the bigger model can be matched by inferring a large number of instances using the smaller model and only a few instances using the bigger model.
Thus, by carefully deciding when to use bigger/more expensive models, the computational efficiency of NLP systems can be improved.
\textit{So, how should we decide which model(s) to use for a given test instance?}
Figure \ref{fig:teaser} illustrates an approach to achieve this; it infers an instance sequentially through models (ordered in increasing order of capacity) and uses a threshold over the maximum softmax probability (\textit{MaxProb}) to decide whether to output the prediction or pass it to the next model in sequence.
% Thus, instances that are predicted with high MaxProb get answered at early stages as their predictions are likely to be correct and the remaining ones get passed to the larger models.
The intuition behind this approach is that MaxProb shows a positive correlation with predictive correctness. 
Thus, instances that are predicted with high MaxProb get answered at early stages as their predictions are likely to be correct and the remaining ones get passed to the larger models.
Hence, by avoiding inference through large and expensive models (primarily for easy instances), cascading makes the system computationally efficient while maintaining high prediction performance.
% is more likely to be predicted correctly. 
% In the figure, given input is first inferred using the smallest model, this model fails to predict it with sufficient confidence therefore it is passed to a larger capacity model.
% The larger model's prediction confidence exceeds the threshold and the system outputs its final prediction.
% We explore several such cascading methods (Section \ref{subsec_approaches}) and compare their performance in multiple experimental settings (Section \ref{sec_experiments}).

% Another advantage of cascading is that it allows custom computation costs as different number of models can be used for inference.
% For example, in case of cascading with \textit{MaxProb}, varying the confidence thresholds leads to different computation costs for the system.
% In contrast, using a single model does not provide such flexibility as it has a fixed inference cost.
We describe several such cascading methods in Section \ref{subsec_approaches}.
Furthermore, cascading allows custom computation costs as different number of models can be used for inference.
We compute accuracies for a range of costs and plot an accuracy-cost curve.
Then, we calculate its area (AUC) to quantify the efficacy of the cascading method.
Larger the AUC value, the better the method is as it implies higher accuracy on average across computation costs.

% We explore several cascading methods  and compute their accuracies for a range of computation costs to plot an accuracy-cost curve for each method.
% We compute accuracy of cascading systems for a range of computation costs and plot accuracy-cost curves.
% Then, we calculate area under these curves (AUC) to estimate the efficacy of different cascading methods.
% We describe several cascading methods (Section \ref{subsec_approaches}) and compare their performance in multiple experimental settings.

We conduct comprehensive experiments with $10$ diverse NLU datasets in multiple task settings that differ in the number of models available for cascading ($K$ value from Section \ref{sec_cascading}).
We first demonstrate that cascading achieves considerable improvement in computational efficiency.
% Figure \ref{fig: overall_improvement_K_3} (left) illustrates efficiency improvements achieved by a cascading method over the largest model ($M_3$) in K=3 setting.
For example, in case of QQP dataset, cascading system achieves $88.93\%$ computation improvement over the largest model ($M_3$) in K=3 setting i.e. it requires just $11.07\%$ of the computation cost of model $M_3$ to attain equal accuracy.
Then, we show that cascading also achieves improvement in prediction accuracy.
% Figure \ref{fig: overall_improvement_K_3} (right) illustrates the accuracy improvements achieved over $M_3$ in K=3 setting.
For example, on CB dataset, the cascading system achieves $2.18\%$ accuracy improvement over $M_3$ in the K=3 setting.
Similar improvements are observed in settings with different values of $K$.
% We note that cascading primarily improves computational efficiency while accuracy improvement is just a by-product.
% We conduct a thorough analysis and justify reasons behind these improvements.
Lastly, we show that introducing additional model in the cascade further increases the efficiency benefits.
% show benefits of cascading in out-of-domain (OOD) settings.
% accuracy improvement is just a by-product of cascading as efficiency improvement is its primary achievement.
% Finally, we show such benefits of cascading in multiple task settings for a number of datasets.

% achieves up to $88.93\%$ efficiency improvement i.e. the cascading system requires just $11.07\%$ computation of model $M_3$ to match its accuracy;
% for $K=2$, 
% Specifically, in the setting with two models ($K=2$), we show that cascading requires as low as $0.32X$ computation cost to match the accuracy of the large model ($M_2$).

In summary, our contributions and findings are:
% as follows:
\begin{enumerate}[noitemsep,nosep,leftmargin=*]
    \item \textbf{Model Cascading:} We provide mathematical formulation of model cascading, explore several methods, and systematically study its benefits.
    % To the best of our knowledge, we are the first to systematically study this in NLP.
    
    \item \textbf{Cascading Improves Efficiency}: 
    Using accuracy-cost curves, we show that cascading systems require much lesser computation cost to attain accuracies equal to that of big models.
    % For example, in the setting with two models ($K=2$), cascading with MaxProb requires as low as $0.25X$ computation cost to match the accuracy of the large model ($M_2$).
    % From the accuracy-cost curves, we observe that the accuracy of the system usually increases with the increase in computation cost. This is expected as the system gets to utilize larger (and often more accurate) models for more number of instances. On this curve, we identify the computation cost at which the cascading system's accuracy is equal to the large model's accuracy and show that this cost is considerably less than the cost of the large model.
    % we show that cascading systems could match the accuracy of individual models at much lesser computation cost.
    % For instance, in a setting with two models ($K=2$), the system with MaxProb cascading matches the accuracy of the larger model ($M_2$) at as low as $0.25X$ of its computation cost.
    
    \item \textbf{Cascading Improves Accuracy: } 
    We show that cascading systems consistently achieve superior prediction performance than even the largest model available in the task setting. 
    % We further demonstrate both the efficiency and accuracy benefits of cascading in OOD settings.
    % We show that cascading systems even surpass the accuracy of all individual models in the collection. 
    % With increase in computation cost, the accuracy of a cascading system usually also increases. 
    % We show that cascading achieves higher accuracies than the individual models at equivalent computation costs.
    % For example, in $K=2$ setting, MaxProb achieves up to $x\%$ higher accuracy than the model $M_2$.
    % they achieves higher accuracy than the individual models at equivalent computation costs.
    % Thus, beyond the computation cost found in the previous point (2), the accuracy of cascading system surpasses the accuracy of all individual models in the collection.
    % Cascading systems start to achieve higher accuracy than the large model from the computation cost identified in the previous point.
    % We also compare the accuracy achieved by the cascading system with that of the large model at its computation cost.

    \item \textbf{Comparison of Cascading Methods: }
    We compare performance of our proposed cascading methods and find that \textit{DTU} (\ref{subsec_approaches}) outperforms all others by achieving the highest AUC of accuracy-cost curves on average. 
    
    % In $K=2$ setting, MaxProb and DTU achieve nearly the same AUC but in $K=3$ setting, DTU clearly outperforms all methods.
    
\end{enumerate}

% Smaller variants have been proposed such as bert-mini, bert-medium etc. but their accuracy is poor.
% There is a variety of adaptive/dynamic inference approaches proposed, however, a general down-side for many of these methods is that often times they require a careful architecture design, manipulation of network modules, or even re-training.

% We release our code with experimental setup and hope our work will encourage further research in `model cascading', a simple yet effective technique to jointly improve efficiency and accuracy.

We note that model cascading is trivially easy to implement, can be applied to a variety of problems, and can have good practical values.

% leverages multiple models of different capacities to jointly improve accuracy and efficiency of NLP systems.

% \subsection{Improving Accuracy}

% \paragraph{Large Pre-trained Models:}
% With the success of transformer-based pre-trained language models such as BERT \cite{devlin-etal-2019-bert}, a trend in NLP has been to develop even larger pre-trained language models such as RoBERTa \cite{Liu2019RoBERTaAR}, GPT series \cite{radford2019language,NEURIPS2020_1457c0d6}, and T5 \cite{JMLR:v21:20-074} to push the performance to new levels.

% \paragraph{Model Ensembling: } 
% In ensembling, the system aggregates predictions from multiple models to achieve better performance. 
% Ensembling has been used is several recent works \cite{}.

% Model ensembling is computationally very expensive as all the models need to make an inference.
% Though these approaches improve the accuracy, they are computationally very expensive.
% This is because with the increase in number of model parameters the computation cost increases and using multiple models also increases the cost.
% The trend in NLP has been to develop ever larger models to achieve better accuracy

\section{Related Work}
\label{subsec_related_work}
% A recent trend in NLP research has been to develop large-scale pre-trained language models \cite{radford2019language,NEURIPS2020_1457c0d6,JMLR:v21:20-074}. These models achieve impressive performance but are computationally very expensive. 
% To this end, improving efficiency of NLP systems has been a focus of active research and resulted in development of several techniques, such as

% (1) \textit{network pruning} that progressively removes model weights from a big network resulting in a smaller model \cite{wang-etal-2020-structured,guo-etal-2021-parameter}, 
% % (2) \textit{quantization} that reduces the number of unique values required to represent model weights and activations \cite{shen2020q,zhang-etal-2020-ternarybert},
% (2) \textit{early exiting} that allows multiple exist paths in a model \cite{xin-etal-2020-deebert},
% (3) \textit{adaptive inference} that adjusts the model size by adaptively selecting width and depth,
% (4) \textit{knowledge distillation} that transfers `dark-knowledge' from a large teacher model to a shallow student model \cite{jiao-etal-2020-tinybert,li2022dq}, and
% (5) \textit{input reduction} that eliminates less contributing tokens from the input text to speed up inference

In recent times, several techniques have been developed to improve the efficiency of NLP systems, such as 
\textit{network pruning} \cite{wang-etal-2020-structured,guo-etal-2021-parameter,NEURIPS2020_b6af2c97}, 
\textit{quantization} \cite{shen2020q,zhang-etal-2020-ternarybert,tao2022compression},
\textit{knowledge distillation} \cite{clark-etal-2019-bam,jiao-etal-2020-tinybert,li-etal-2022-dq,mirzadeh2020improved}, and
\textit{input reduction} \cite{modarressi-etal-2022-adapler}.
Our work is more closely related to dynamic inference \cite{xin-etal-2020-deebert} and adaptive model size \cite{goyal2020power,kim-cho-2021-length,NEURIPS2020_6f5216f8,soldaini-moschitti-2020-cascade}.
% closest but non-trivially different from ours are:

% \paragraph{Early Existing: }
\citet{xin-etal-2020-deebert} proposed Dynamic early exiting for BERT (DeeBERT) that speeds up BERT inference by inserting extra classification layers between each transformer layer.
It allows an instance to choose conditional exit from multiple exit paths.
% Early exiting saves computation cost.
All the weights (including newly introduced classification layers) are jointly learnt during training.

% \paragraph{Adaptive Size of Model: }
\citet{goyal2020power} proposed Progressive Word-vector Elimination (PoWER-BERT) that reduces intermediate vectors computed along the encoder pipeline.
They eliminate vectors based on significance computed using self-attention mechanism.
\citet{kim-cho-2021-length} extended PoWER-BERT to Length-Adaptive Transformer which adaptively determines the sequence length at each layer.
\citet{NEURIPS2020_6f5216f8} proposed a dynamic BERT model (DynaBERT) that adjusts the size of the model by selecting adaptive width and depth.
They first train a width-adaptive BERT and then distill knowledge from full-size models to small sub-models.

Lastly, cascading has been studied in machine learning and vision with approaches such as Haar-cascade \cite{soo2014object} but is underexplored in NLP.
We further note that cascading is non-trivially different from `ensembling' as ensembling always uses all the available models instead of carefully selecting one or more models for inference. 
% In ensembling, the focus is on accuracy and efficiency is ignored.

Our work is different from existing methods in the following aspects:
(1) Existing methods typically require architectural changes, network manipulation, saliency quantification, knowledge distillation, or complex training procedures. In contrast, cascading is a simple technique that is easy to implement and does not require such modifications,
(2) The computational efficiency in existing methods often comes with a compromise on accuracy. Contrary to this, we show that model cascading surpasses the accuracy of even the largest models,
(3) Existing methods typically require training a separate model for each computation budget; on the other hand, a single cascading system can be adjusted to meet all the computation constraints.
(4) Finally, cascading does not require an instance to be passed sequentially through the model layers; approaches such as \textit{routing} (section \ref{sec_cascading}) allow passing it directly to a suitable model.

\section{Model Cascading}
\label{sec_cascading}
We define model cascading as follows:

\textit{Given a collection of models of varying capacities, the system needs to leverage one or more models in a computationally efficient way to output accurate predictions.}
% output its prediction by leveraging one or more models in an accurate yet computationally efficient way.}

As previously mentioned, a system using only the smallest model would be highly efficient but poor in accuracy and a system using all the available models would be fairly accurate but expensive in computation.
The goal of cascading is to achieve high prediction accuracy while efficiently leveraging the available models.
The remainder of this section is organized as follows:
we provide mathematical formulation of cascading in \ref{subsec_formulation} and
% define its performance metric in \ref{subsec_evaluation}, 
describe its various approaches in \ref{subsec_approaches}.
% without compromising the accuracy.
% Model cascading task requires balancing accuracy and efficiency trade-offs.
% using only the smallest model would result in a system that is highly efficient but poor in prediction performance;
% be the most efficient strategy but will usually have poor accuracy; 
% on the other hand, using all the models would result in a fairly accurate system but computationally very expensive.

\subsection{Formulation}
\label{subsec_formulation}
% \Neeraj{Cost depends on the length, change the formulation accordingly}
Consider a collection of $K$ trained models ($M_1,...,M_K$) ordered in increasing order of their computation cost i.e. for an instance $x$, $c_j^x < c_k^x$ ($\forall$ $j<k$) where $c$ corresponds to the cost of inference.
% their respective computation cost of inference is ($c_1^x,...,c_K^x$) for an input $x$ such that $c_j^x < c_k^x$ ($\forall$ $j<k$).
% Consider a collection of $K$ models ($M_1,...,M_K$) (with respective cost of inference ($c_1,...,c_K$)) ordered in increasing order of their capacity/size i.e. $c_i > c_j$ (for all $i>j$).
The system needs to output a prediction for each instance of the evaluation dataset $D$ leveraging one or more models. 
Let $M_j^{x}$ be a function that indicates whether model $M_j$ is used by the system to make inference for the instance $x$ i.e. 
\begin{equation*}
        M_j^{x} =
            \begin{cases}
              1, & \text{if model $M_j$ is used for instance $x$} \\
              0, & \text{otherwise}
            \end{cases}
\end{equation*}
% Note that the computation cost of inference depends on the sequence length of the input. 
Thus, the average cost of the system for the entire evaluation dataset $D$ is calculated as:
\begin{equation*}
    Cost_{D} = \frac{\sum_{x_i \in D} \sum_{j=1}^{K} M_j^{x_i}\times c_j^{x_i}}
    {|D|} 
\end{equation*}
In addition to this cost, we also measure accuracy i.e. the percentage of correct predictions by the system. 
The goal is to achieve high prediction accuracy while being computationally efficient.
 
% Thus, the computation cost varies with the 
% We plot accuracy-computation cost values for a range of computation costs.

% also compare the cascading accuracy when the computational cost is equivalent to the models with the models' accuracy.
% Similarly, we compare the cascading cost when the accuracy is equivalent to the models.

\paragraph{Performance Evaluation:}
% \label{subsec_evaluation}
With the increase in the computation cost, the accuracy usually also increases as the system leverages large models (that are often more accurate) for more number of instances. 
To quantify the performance of a cascading method, we first plot its accuracy-cost curve by varying the computation costs and then calculate the area under this curve (AUC).
% For each cascading method, we collect accuracy values for a range of computation costs and plot their respective accuracy-cost curve.
% To quantify the performance of a cascading method, we calcualte the area under this curve (AUC).
\textbf{Larger the AUC value, the better the cascading method is} as it implies higher accuracy on average across all computation costs.
We note that the computation cost of the cascading system can be varied by adjusting the confidence thresholds of models in the cascade (described in the next subsection).

Along with the AUC metric, we evaluate efficacy of cascading on two additional parameters: 
\begin{enumerate}[noitemsep,nosep,leftmargin=*]
    \item \textbf{Comparing computation cost of the cascading system at accuracies achieved by each individual model of cascade:} Consider a setting in which the model $M_2$ achieves accuracy $a_2$ at computation cost $c_2$; from the accuracy-cost curve of the cascading system, we compare $c_2$ with the cost of the cascading system when its accuracy is $a_2$.
    
    \item \textbf{Comparing the maximum accuracy of the cascading system with that of the largest model of collection:} We compare accuracy of the largest individual model with the maximum accuracy achieved by the cascading system.
    
\end{enumerate}
Note that the first parameter corresponds to the point of intersection obtained by drawing a horizontal line from accuracy-cost point of each individual model on the accuracy-cost curve.
% and a vertical line (for parameter two).
Refer to the red dashed lines in Figure \ref{fig:acc_cost_curves_K_2} and \ref{fig:acc_cost_curves_K_3} for illustration.
For a cascading system to perform better than the individual models in the cascade, it should have a lower computation cost (in parameter one) and a higher accuracy (in parameter two).

\subsection{Approaches}
\label{subsec_approaches}

We explore the following approaches of selecting which model(s) to use for inference.

\paragraph{Maximum Softmax Probability (MaxProb):}
Usually, the last layer of a model has a softmax activation function that distributes its prediction probability $P(y)$ over all possible answer candidates $Y$. 
% In case of classification tasks, $Y$ is the set of labels.
% For instance, in natural language inference task, $Y$ is \{Entailment, Contradiction, Neutral\}.
% Softmax activation function distributes a model's prediction probability for an instance $x$ over its answer space $Y$.
MaxProb corresponds to the maximum softmax probability assigned by the model i.e.
\begin{equation*}
    MaxProb = \max_{y \in Y} P(y)
\end{equation*}
MaxProb (often termed as prediction confidence) has been shown to be positively correlated with predictive correctness \cite{hendrycks17baseline,hendrycks-etal-2020-pretrained, varshney-etal-2022-investigating} i.e. a high MaxProb value implies a high likelihood for the model's prediction to be correct.
We leverage this characteristic of MaxProb in our first cascading approach. 
Specifically, we infer the given input instance sequentially through the models starting with $M_1$ and use a confidence threshold over MaxProb value to decide whether to output the prediction or pass the instance to the next model in sequence. 

Consider an instance $x$ for which the models till $M_{z-1}$ fail to surpass their confidence thresholds and $M_{z}$ exceeds its threshold then:
% the model $M_z$'s prediction confidence exceeds the threshold then
\begin{equation*}
        M_j^{x} =
            \begin{cases}
              1, & \text{if $j \leq  z$} \\
              0, & \text{if $j > z$}
            \end{cases}
\end{equation*}
The confidence thresholds could be different at different stages.
Figure \ref{fig:teaser} illustrates this approach.

It provides efficiency benefits as it avoids passing easy instances (that can be potentially answered correctly by low-compute models) to the computationally expensive models.
Furthermore, it does not sacrifice the accuracy of system because the difficult instances would often end up being answered by the large (and more accurate) models.
We note that this approach requires additional computation for comparing MaxProb values with thresholds but its cost is negligible in comparison to the cost of model inferences and hence ignored in the overall cost calculation.

% the decision logic in this approach adds an additional negligible computation overhead in comparison to cost of model inferences and hence 

% Specifically, we first pass the instance through the smallest model ($M_1$) and if its MaxProb is above a certain threshold then we use its prediction as the system's output otherwise we pass the instance to the next model i.e. $M_2$. This process is repeated until the MaxProb value exceeds the threshold or all the models have been utilized. 

% Try different approaches - just rely on the final model, take ensemble of the used models.
% MaxProb has been shown to be positively correlated with correctness \cite{} i.e. if a model's MaxProb on two instances is $m_1$ and $m_2$ with $m_1 > m_2$ then the model's prediction on the former instance is more likely to be correct than the latter.

% \paragraph{MaxProb Minus Second Highest Softmax Probability (MmSH):}

\paragraph{Distance To Uniform Distribution (DTU): }
In this approach, we use the distance between the model's softmax probability distribution and the uniform probability distribution as the confidence estimate (in place of MaxProb) to decide whether to output the prediction or pass the instance to the next model in sequence i.e. 
\begin{equation*}
    DTU = ||P(Y)-U(Y)||_2
\end{equation*}
where $U(Y)$ corresponds to the uniform output distribution. For example, in case of a task with $4$ classification labels, $U(Y) = [0.25, 0.25, 0.25, 0.25]$.
The intuition behind this approach is to leverage the entire shape of the output probability distribution and not just the highest probability as in MaxProb.

% \paragraph{Energy: }
% Here, we use temperature-scaled energy \cite{lecun2006tutorial} as confidence estimate i.e. 
% % \begin{equation*}
% %     Energy = -log\sum
% % \end{equation*}
% \[
%      Energy = - T*log{ \sum_{i=1}^{C} e^{p(y_i)/T}}
% \]
% where $C$ is the number of classes in the task and $T$ is temperature.

\paragraph{Random:}

In this approach, instead of using a metric such as MaxProb or DTU to decide which instances to pass to the next model in sequence, we do this instance selection process at random.
This serves as a baseline cascading method.
% We compute accuracy-cost pairs for each possible proportion and plot its accuracy-cost curve.

% In this approach, we divide the evaluation dataset among the available models in each possible proportion, and compute accuracy and cost values for each scenario.
% Using these accuracy-cost pairs, we plot its accuracy-cost curve.
% Table \ref{tab:random_approach_illustration} illustrates this procedure for $K=2$ setting.
% Here, only one model is used for each instance (either $M_1$ or $M_2$) unlike previous approaches where multiple models were used.
% However, multiple models can also be used randomly but that would result in increase in computation cost and we show that even the low-compute Random approach fails to outperform other approaches.
% \input{Tables/random_approach_illustration}

\paragraph{Heuristic:}
Here, we use a heuristic derived from the input text to decide which instances to pass to the next model in sequence. 
Specifically, we use length of the input text as the heuristic.

\paragraph{Routing:}
In this approach, instead of sequentially passing an instance to bigger and bigger models, we skip intermediate models and pass the instance directly to a suitable model based on its maxProb value. 
For example, in $K=3$ setting, we first infer using $M_1$ and if its maxProb is very low then we skip $M_2$ and directly pass it to $M_3$. 
On the other hand, if its maxProb is sufficiently high (but below $M_1$'s output threshold) then we pass it to $M_2$.
The intuition behind this approach is that the system might save inference cost of intermediate models by directly using a suitable model that is likely to answer it correctly.
This approach is not applicable for $K=2$ as there is only one option to route after inference through the model $M_1$.

% not necessarily sequentially, increasing capacity, 

% Model Cascading has been studied in Machine Learning. 

\section{Experiments}
\label{sec_experiments}
\subsection{Experimental Details}
\paragraph{Datasets: }
% We conduct comprehensive experiments with $10$ NLU datasets spanning over several tasks, such as natural language inference, duplicate detection, and sentiment classification.
We experiment with a diverse set of NLU classification datasets:
SNLI~\cite{bowman-etal-2015-large},
Multi-NLI~\cite{williams-etal-2018-broad}, 
Dialogue-NLI~\cite{welleck-etal-2019-dialogue},
Question-NLI \cite{wang-etal-2018-glue},
QQP \cite{iyer2017first}, 
MRPC \cite{dolan2005automatically}, 
PAWS \cite{zhang-etal-2019-paws},
SST-2 \cite{socher-etal-2013-recursive}, 
COLA \cite{warstadt-etal-2019-neural}, and
CommitmentBank \cite{Marneffe2019TheCI}.

\paragraph{Models: }

We use the following variants of BERT \cite{devlin-etal-2019-bert}:
BERT-mini (11.3M parameters),
BERT-medium (41.7M parameters),
BERT-base (110M parameters), and
BERT-large (340M parameters) for our experiments. 
Table \ref{tab:stats} shows the computation cost (in FLOPs) of these models for different input text sequence lengths.
We use sequence length of $50$ for COLA, $80$ for SST2, $100$ for QQP, $120$ for MNLI, DNLI, SNLI, $150$ for QNLI, MRPC, PAWS, and $275$ for CommitmentBank datasets following the standard experimental practice.
% \paragraph{Training Details: }
We run all our experiments on Nvidia V100 GPUs with a batch size of $32$ and learning rate ranging in $\{1{-}5\}e{-}5$. 

% In next subsection (\ref{subsection_k_2}), we study the efficacy of cascading in K=2 setting and then in \ref{subsec_K_3}, we investigate the impact of introducing another model in the setup (making it K=3).

In the following subsections, we study the effect of cascading in multiple settings that differ in the number of models in the cascade i.e. $K$ value in the task formulation.

\begin{figure*}[t]
\centering

    \begin{subfigure}{.28\linewidth}
        \includegraphics[width=\linewidth]{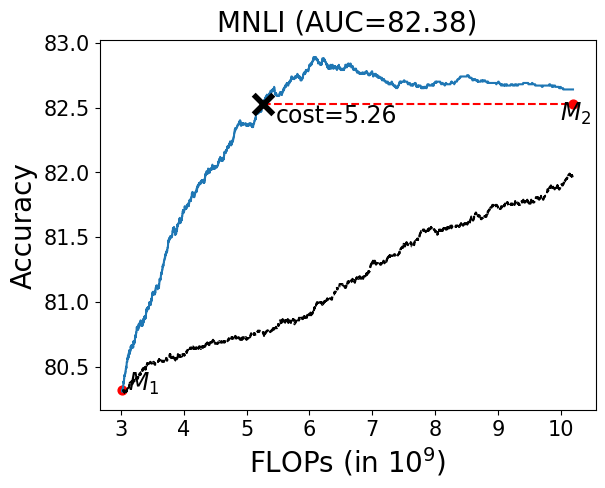}
    \end{subfigure}
    % \begin{subfigure}{.3\linewidth}
    %      \includegraphics[width=\linewidth]{Pictures/K_2/maxProb/medium_base/qnli.png}
    % \end{subfigure}
    \begin{subfigure}{.28\linewidth}
         \includegraphics[width=\linewidth]{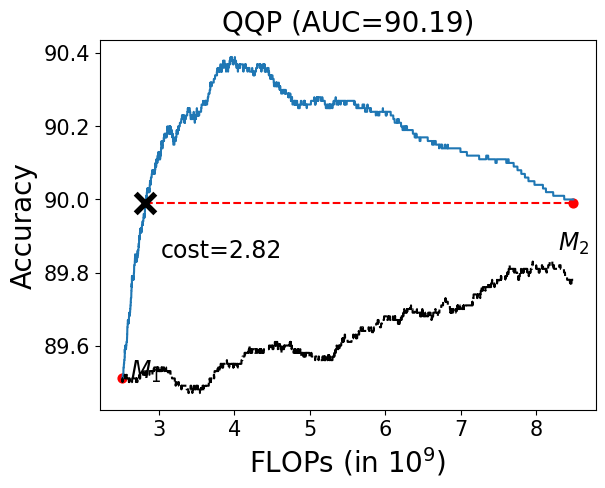}
    \end{subfigure}
    \begin{subfigure}{.28\linewidth}
         \includegraphics[width=\linewidth]{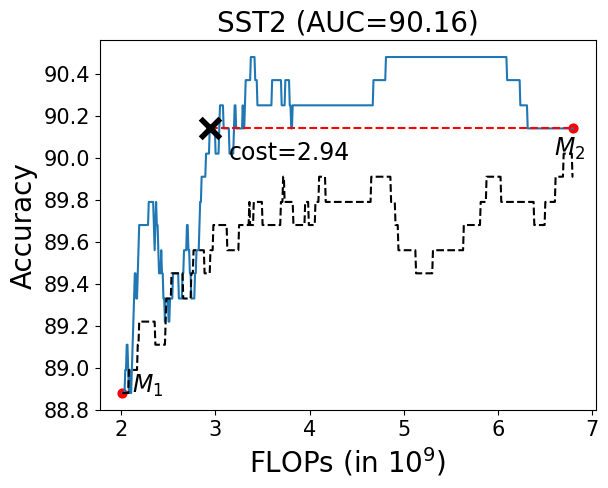}
    \end{subfigure}
    % \begin{subfigure}{.3\linewidth}
    %     \includegraphics[width=\linewidth]{Pictures/K_2/maxProb/medium_base/cola.png}
    % \end{subfigure}
    % \begin{subfigure}{.3\linewidth}
    %      \includegraphics[width=\linewidth]{Pictures/K_2/maxProb/medium_base/commitmentbank.png}
    % \end{subfigure}
    
    % \begin{subfigure}{.3\linewidth}
    %      \includegraphics[width=\linewidth]{Pictures/K_2/maxProb/medium_base/dnli.png}
    % \end{subfigure}
    \begin{subfigure}{.28\linewidth}
         \includegraphics[width=\linewidth]{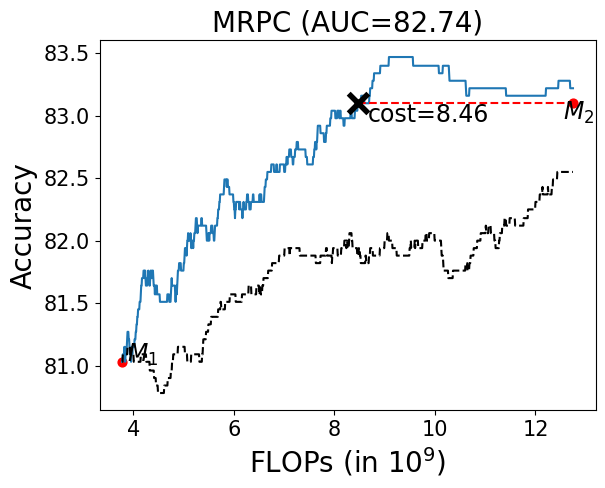}
    \end{subfigure}
    \begin{subfigure}{.28\linewidth}
         \includegraphics[width=\linewidth]{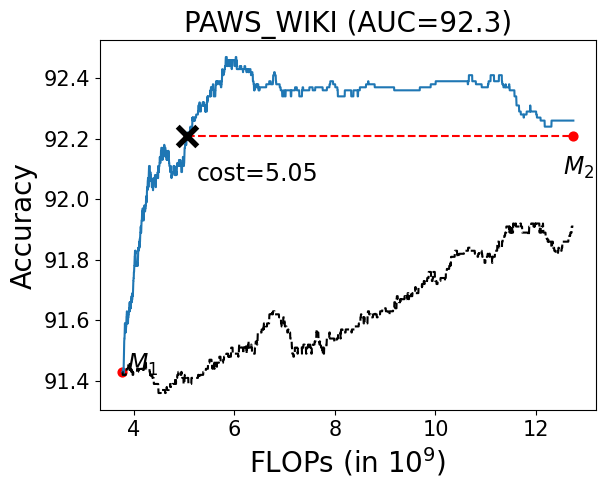}
    \end{subfigure}
    \begin{subfigure}{.28\linewidth}
         \includegraphics[width=\linewidth]{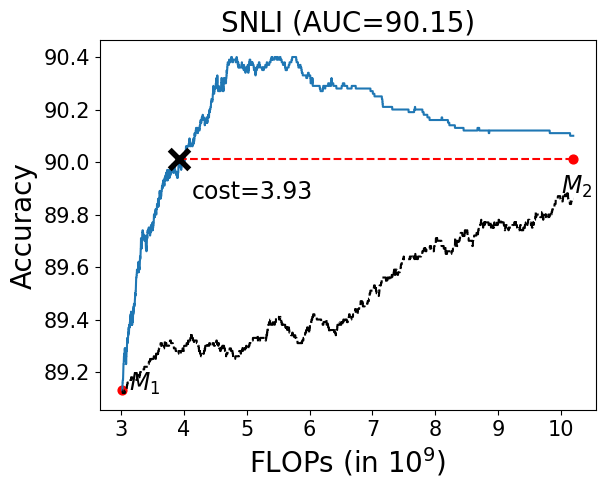}
    \end{subfigure}
    \caption{Accuracy-computation cost curves for cascading with \textit{MaxProb} (in blue) and \textit{Random} baseline (in black) methods in K=2 setting. Red points correspond to the accuracy-cost values of individual models $M_1$ and $M_2$. Points of intersection of red dashed lines drawn from $M_2$ on the blue curve correspond to the evaluation parameters described in Section \ref{sec_cascading}. \textit{MaxProb} outperforms \textit{Random} baseline as it achieves considerably higher AUC.}
    \label{fig:acc_cost_curves_K_2}    
\end{figure*}

\begin{table}[t]
    \centering
    \small
    \begin{tabular}{@{}l|cccc@{}}
        \toprule
         & \textbf{Mini} & \textbf{Medium} & \textbf{Base} & \textbf{Large}\\
        \textbf{Length} & (128M) & (474M) & (1.3G) & (3.8G)\\
        \midrule
        % 50 & 11.11 & 11.11  & 11.11  & 11.11  \\
        % 80 & 11.11 & 11.11  & 11.11  & 11.11  \\
        % 100 & 11.11 & 11.11  & 11.11  & 11.11  \\
        % 120 & 11.11 & 11.11  & 11.11  & 11.11  \\
        % 150 & 11.11 & 11.11  & 11.11  & 11.11  \\
        % 220 & 11.11 & 11.11  & 11.11  & 11.11  \\
        % 275 & 11.11 & 11.11  & 11.11  & 11.11  \\

        50 & 0.16 & 1.26 & 4.25 & 5.10 \\ 
        80 & 0.25 & 2.01 & 6.80 & 24.16 \\ 
        100 & 0.31 & 2.52 & 8.49 & 30.20 \\ 
        120 & 0.38 & 3.02 & 10.19 & 36.24 \\
        150 & 0.47 & 3.78 & 12.74 & 45.30 \\
        220 & 0.69 & 5.54 & 18.69 & 66.44 \\
        275 & 0.87 & 6.92 & 23.36 & 83.05 \\
        
    \bottomrule
    \end{tabular}
    \caption{Inference cost (in $10^{9}$ FLOPs) of BERT variants for different input text sequence lengths. We also specify the storage size of the models in this table.}
    \label{tab:stats}
\end{table}
\subsection{Cascading with Two Models (K=2)}
\label{subsection_k_2}
\begin{table*}[]
\small
\resizebox{\textwidth}{!}{%
\begin{tabular}{l|llllllllll}
% \begin{tabular}{ll|lllllllllll}
\toprule
% \multicolumn{1}{c}{\multirow{1}{*}{\textbf{Model}}} &
%   \multicolumn{1}{c}{\multirow{1}{*}{\textbf{Metric}}}  &
%   \multicolumn{1}{c}{\textbf{MNLI}} &
%   \multicolumn{1}{c}{\textbf{MNLI}} &
%   \multicolumn{1}{c}{\textbf{MNLI}} &
%   \multicolumn{1}{c}{\textbf{MNLI}} &
%   \multicolumn{1}{c}{\textbf{MNLI}} &
%   \multicolumn{1}{c}{\textbf{MNLI}} &
%   \multicolumn{1}{c}{\textbf{MNLI}} &
%   \multicolumn{1}{c}{\textbf{MNLI}} &
%   \multicolumn{1}{c}{\textbf{MNLI}} &
%   \multicolumn{1}{c}{\textbf{MNLI}}

\textbf{Method} &
\textbf{MNLI} &  \textbf{QNLI} &  \textbf{QQP} &  \textbf{SST2} &  \textbf{COLA} &  \textbf{CB} &  \textbf{DNLI} &  \textbf{MRPC} &  \textbf{PAWS} &  \textbf{SNLI} \\

\midrule
    
    % Random & 
    % 81.48 &	84.71 &		89.72 &		89.86 &	 78.81 &		78.51 &	 85.51 &		82.11 &	 91.79 &		89.65 \\
    
    Random & 
    81.16 & 	84.24 & 		89.64 & 		89.64 & 	 77.97 & 		77.27 & 	 85.35 & 		81.76  & 	91.63 & 		89.50 \\
    
    Heuristic & 
    81.13 & 	84.15 & 	89.69 & 	89.06  &  79.45 & 	76.79 &  85.30 & 	81.54 &  91.65 & 	89.45 \\
    
    \midrule
    
    MaxProb & 
    82.38 &	85.73 &		90.19 &		90.16 &	 80.25 &		80.40 &	 85.27 &		82.74 &	 92.30 &		90.15 \\
    
    DTU &
    82.38 &	85.73 &		90.18 &		90.16 &	 80.25 &		80.70 &	 85.24 &		82.74 &	 92.30 &		90.16 \\

\bottomrule
\end{tabular}
}
\caption{Comparing AUC values of different cascading methods in K=2 setting. \textit{Random} and \textit{Heuristic} correspond to the cascading baselines. \textit{MaxProb} and \textit{DTU} outperform both the baselines.}
\label{tab:auc_K_2}

\end{table*}

\subsubsection{Problem Setup}
In this setting, we consider two trained models BERT-medium (41.7M parameters) as $M_1$ and BERT-base (110M parameters) as $M_2$.
% Following the task formulation, $M_2$ has more parameters and hence higher inference cost than $M_1$.
We analyze results for other model combinations (such as medium, large and mini, large) in Appendix \ref{supp_sec_k_2}.

% more parameters and hence higher cost than M1.

\subsubsection{Results}
Recall that the computation cost of a cascading system can be controlled by changing the $M_j$ values.
For example, in case of MaxProb, changing the confidence threshold value would result in different $M_j$ values and hence different cost and accuracy values. 
% For each dataset, we plot accuracy-cost curve of cascading methods and also show the accuracy-cost points for models M1 and M2 in the same figure.
Figure \ref{fig:acc_cost_curves_K_2} shows accuracy-cost curves for two cascading approaches: MaxProb (in blue) and Random Baseline (in black).
In the same figure, we also show accuracy-cost points for the individual models $M_1$ and $M_2$.
To avoid cluttering these figures, we plot accuracy-cost curves for other approaches in separate figures and present them in Appendix \ref{supp_sec_k_2}.
However, to compare the performance of these methods, we provide their AUC values (of their respective accuracy-cost curves) in Table \ref{tab:auc_K_2}.

\paragraph{Efficiency Improvement:}
The accuracy-cost curves show that the cascading system matches the accuracy of the larger model $M_2$ at considerably lesser computation cost.
This cost value corresponds to the point of intersection on the curve with a straight horizontal line drawn from $M_2$ (red dashed line).
For example, in case of QQP, model $M_2$ achieves $89.99\%$ accuracy at average computation cost of $8.49$ $\times$ $10^{9}$ FLOPs while the cascading system achieves the same accuracy at only $2.82$ $\times$ $10^{9}$ FLOPs. Similarly, in case of MNLI, $M_2$ achieves $82.53\%$ accuracy at cost of $10.19$ $\times$ $10^{9}$ FLOPs while the cascading system achieves the same accuracy at only $5.26$ $\times$ $10^{9}$ FLOPs.
Such improvements are observed for all datasets.
This efficiency benefit comes from using the smaller models for a large number of instances and passing only a few instances to the larger models.
% On average, cascading achieves $X$ efficiency improvement across all datasets.

\paragraph{Accuracy Improvement:}
% Accuracy-cost curves show that the accuracy usually increases with the increase in the computation cost. 
From the accuracy-cost curves, it can be observed that beyond the cost value identified in the previous paragraph (where the red dashed line intersects the accuracy-cost curve), the cascading system outperforms model $M_2$ in terms of accuracy.
For example, in case of QQP, cascading with MaxProb achieves accuracy of up to $90.39\%$ that is higher than the accuracy of $M_2$ ($89.99\%$). 
Similar improvements are observed for all other datasets.
% in case of MNLI, cascading with MaxProb achieves $82.89\%$ accuracy that is higher than the accuracy of both models $M_1$ ($80.32\%$) and $M_2$ ($82.53\%$). 
We note that the accuracy improvement is a by-product of cascading, its primary benefit remains to be the improvement in computational efficiency.
% primarily it provides efficiency benefits while maintaining the accuracy.

Higher accuracy achieved by the cascading system (that uses $M_1$ for some instances and conditionally also uses $M_2$ for others) than the larger model $M_2$ implies that $M_1$, despite being smaller in size is more accurate than $M_2$ on at least a few instances. 
Though, on average across all instances, $M_2$ has higher accuracy than $M_1$.
The cascading system uses $M_1$ for instances on which it is sufficiently confident and thus more likely to be correct. Only the instances on which it is not sufficiently confident get passed to the bigger model. 
% Cascading leverages $M_1$ for some of these instances and thus achieves higher overall accuracy.
This supports the findings of recent works such as \cite{zhong-etal-2021-larger,varshney-etal-2022-ildae} that conduct instance-level analysis of models' predictions.
We further analyze these results in the next paragraphs.

\paragraph{Comparing Cascading Approaches:}
Figure \ref{fig:acc_cost_curves_K_2} demonstrates that MaxProb cascading approach clearly outperforms the `Random' cascading baseline.
In Table \ref{fig:acc_cost_curves_K_2}, we compare AUC of respective accuracy-cost curves of various cascading approaches.
Both MaxProb and DTU outperform both the baseline methods (Random and Heuristic).
In $K=2$ setting, both MaxProb and DTU achieve roughly the same performance on average across all datasets.
The gap between MaxProb and DTU becomes more significant in $K=3$ setting (\ref{subsec_K_3}).

\begin{figure}[t!]
    \centering
    \includegraphics[width=6cm]{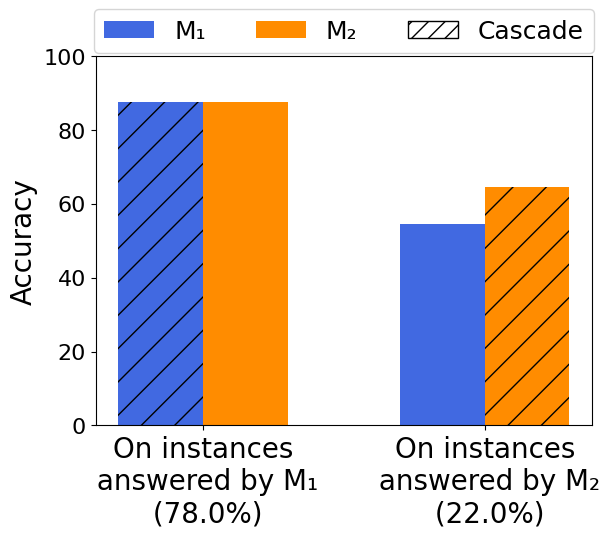}
    \caption{Comparing accuracy of individual models $M_1$ and $M_2$ on the instances answered by each model when used as cascade for MNLI dataset in K=2 setting.}
    \label{fig:contribution_K_2_MNLI}
\end{figure}

\paragraph{Contribution of $M_1$ and $M_2$ in the  Cascade:}
To further analyze the performance of the cascading system, we study the contribution of individual models $M_1$ and $M_2$ in the cascade.
Figure \ref{fig:contribution_K_2_MNLI} shows the contribution of $M_1$ and $M_2$ for MNLI dataset when the cost is $5.26 \times 10^9$ FLOPs i.e. the point at which the accuracy of the cascading system is equal to that of the bigger model $M_2$ (intersection point of the horizontal red dashed line with the accuracy-cost curve of the cascading system in Fig \ref{fig:acc_cost_curves_K_2}).
At this point, the cascade system uses $M_1$ for 78\% instances and $M_2$ for the remaining 22\% instances.
The accuracy of $M_1$ on its 78\% instances (87.6\%) would be equal to that of $M_2$ on those 78\% instances as the overall accuracy of system on complete dataset (100\% instances) is equal to that of $M_2$.
However, this does not imply that the instance-level predictions of the two models on those $78\%$ would be exactly the same. Though, their predictions overlap in majority of the cases.
% We find that their predictions overlap only in $\sim85\%$ cases.

Figure \ref{fig:contribution_K_2_MNLI} also shows that the accuracy of model $M_1$ on the instances that got passed to $M_2$ in the cascade system is significantly lesser (by 33.12\%) than on the instances that $M_1$ answered (blue bars). 
$M_2$ achieves 10.12\% higher accuracy on those instances than $M_1$. Therefore, the cascading system utilizes the models efficiently by using the smaller model $M_1$ for the easy instances and the larger model $M_2$ for the difficult ones.
We analyze these results for other datasets in Appendix \ref{supp_contribution_medium_base}.

% \paragraph{Cascading in OOD:}
% We also study the efficacy of cascading in out-of-domain (OOD) setting for Duplicate Detection datasets.
% Specifically, we evaluate the QQP trained model on other duplicate detection datasets: MRPC and PAWS.
% Figure \ref{fig:ood_K_2} shows the accuracy-cost curves for both the datasets.
% Similar to the IID setting, cascading achieves improvement in both efficiency and accuracy.

% \begin{figure}[t!]
%     \centering
%     \includegraphics[width=6cm]{Pictures/ood/K_2/maxProb/medium_base/qqp/mrpc.png}
%     \caption{Accuracy-cost curves for cascading with MaxProb on OOD datasets.}
%     \label{fig:ood_K_2_MRPC}
% \end{figure}

% \begin{figure}[t!]
%     \centering
%     \begin{subfigure}{0.48\linewidth}
%         \includegraphics[width=\linewidth]{Pictures/ood/K_2/maxProb/medium_base/qqp/mrpc.png}
%     \end{subfigure}
%     \begin{subfigure}{0.48\linewidth}
%          \includegraphics[width=\linewidth]{Pictures/ood/K_2/maxProb/medium_base/qqp/paws_wiki.png}
%     \end{subfigure}
    
%     \caption{Accuracy-cost curves for cascading with \textit{MaxProb} in OOD setting. Models trained on QQP dataset are evaluated on OOD datasets: MRPC and PAWS.}
%     \label{fig:ood_K_2}    
% \end{figure}

\subsection{Cascading with Three Models (K=3)}
\label{subsec_K_3}

\begin{figure*}[t]
\centering

    \begin{subfigure}{.3\linewidth}
        \includegraphics[width=\linewidth]{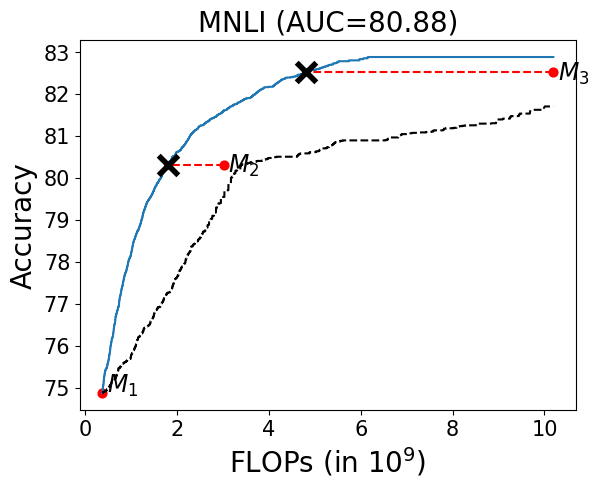}
    \end{subfigure}
    % \begin{subfigure}{.3\linewidth}
    %      \includegraphics[width=\linewidth]{Pictures/K_2/maxProb/medium_base/qnli.png}
    % \end{subfigure}
    \begin{subfigure}{.3\linewidth}
         \includegraphics[width=\linewidth]{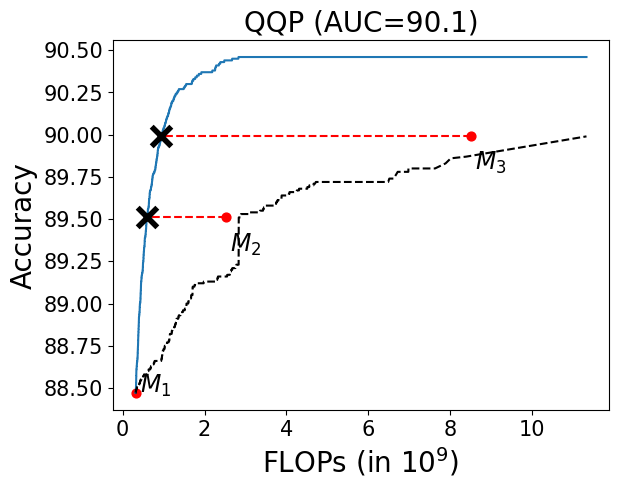}
    \end{subfigure}
    \begin{subfigure}{.3\linewidth}
         \includegraphics[width=\linewidth]{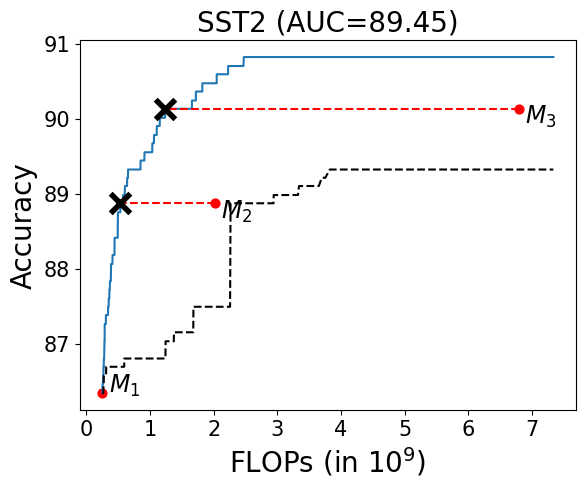}
    \end{subfigure}
    % \begin{subfigure}{.3\linewidth}
    %     \includegraphics[width=\linewidth]{Pictures/K_2/maxProb/medium_base/cola.png}
    % \end{subfigure}
    % \begin{subfigure}{.3\linewidth}
    %      \includegraphics[width=\linewidth]{Pictures/K_2/maxProb/medium_base/commitmentbank.png}
    % \end{subfigure}
    
    % \begin{subfigure}{.3\linewidth}
    %      \includegraphics[width=\linewidth]{Pictures/K_2/maxProb/medium_base/dnli.png}
    % \end{subfigure}
    \begin{subfigure}{.3\linewidth}
         \includegraphics[width=\linewidth]{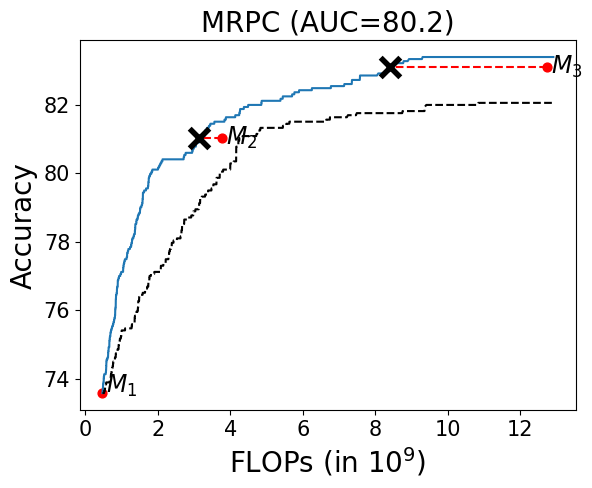}
    \end{subfigure}
    \begin{subfigure}{.3\linewidth}
         \includegraphics[width=\linewidth]{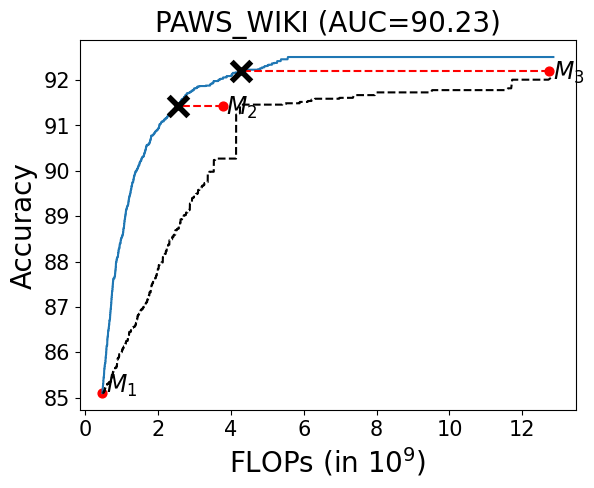}
    \end{subfigure}
    \begin{subfigure}{.3\linewidth}
         \includegraphics[width=\linewidth]{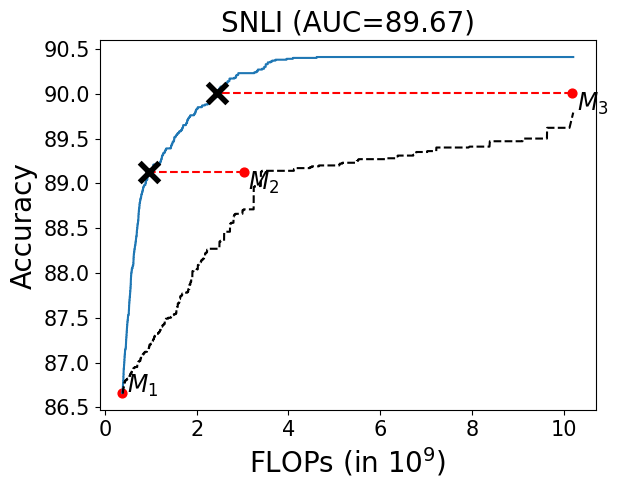}
    \end{subfigure}
    \caption{Accuracy-computation cost curves for cascading with MaxProb (in blue) and Random baseline (in black) methods in K=3 setting. Accuracy-cost values of individual models $M_1$, $M_2$, and $M_3$ are shown in red.
    Note that $M_1$ here is different from $M_1$ in Figure \ref{fig:acc_cost_curves_K_2}.
    \textit{MaxProb} outperforms \textit{Random} baseline as it achieves higher AUC.}
    \label{fig:acc_cost_curves_K_3}    
\end{figure*}
\begin{table*}[]
\small
\resizebox{\textwidth}{!}{%
\begin{tabular}{l|llllllllll}
% \begin{tabular}{ll|lllllllllll}
\toprule
% \multicolumn{1}{c}{\multirow{1}{*}{\textbf{Model}}} &
%   \multicolumn{1}{c}{\multirow{1}{*}{\textbf{Metric}}}  &
%   \multicolumn{1}{c}{\textbf{MNLI}} &
%   \multicolumn{1}{c}{\textbf{MNLI}} &
%   \multicolumn{1}{c}{\textbf{MNLI}} &
%   \multicolumn{1}{c}{\textbf{MNLI}} &
%   \multicolumn{1}{c}{\textbf{MNLI}} &
%   \multicolumn{1}{c}{\textbf{MNLI}} &
%   \multicolumn{1}{c}{\textbf{MNLI}} &
%   \multicolumn{1}{c}{\textbf{MNLI}} &
%   \multicolumn{1}{c}{\textbf{MNLI}} &
%   \multicolumn{1}{c}{\textbf{MNLI}}

\textbf{Method} &
\textbf{MNLI} &  \textbf{QNLI} &  \textbf{QQP} &  \textbf{SST2} &  \textbf{COLA} &  \textbf{CB} &  \textbf{DNLI} &  \textbf{MRPC} &  \textbf{PAWS} &  \textbf{SNLI} \\

\midrule

    Random &
    78.77 & 	80.58 & 	88.97 & 	87.00  & 76.55 & 	77.28 &  84.49 & 	78.30 &  87.74	 & 88.12
    \\
    Heuristic &
    78.85 & 	80.44 & 	88.87 & 	87.67  &  76.28	 & 77.11  & 84.46	 & 77.59  & 88.28	 & 88.27
    \\
    \midrule
    % \hdashline
    % \hline[dashed]
    MaxProb & 
    80.89 & 	82.97 & 		90.1 & 		89.45 & 	  78.66 & 		78.31 & 	 85.17 & 		80.2 & 	 90.23 & 		89.67 
    \\
    
    DTU &
    \textbf{80.98} & 	\textbf{83.28} & 		\textbf{90.15}	 & 	\textbf{89.6} & 	  \textbf{78.87} & 		\textbf{78.52}  & 	\textbf{85.20} & 		80.42  & 	90.46 & 		\textbf{89.72}
    
    \\
    
    Routing &
    80.55 & 	82.93 & 		89.92 & 		89.52 & 	 78.60 & 		74.58  & 	 85.20	 & 	\textbf{80.97}  & 	\textbf{90.68} & 		89.46
    
    \\

\bottomrule
\end{tabular}
}
\caption{Comparing AUC values of different cascading methods in K=3 setting. \textit{Random} and \textit{Heuristic} correspond to the cascading baselines. \textit{DTU} outperforms other cascading methods on average.}
\label{tab:auc_K_3}

\end{table*}

\subsubsection{Problem Setup}
Now, we study the effect of introducing another model in the problem setup of K=2 setting.
Specifically, we consider three models: BERT-mini (11.3M parameters) as $M_1$, BERT-medium (41.7M parameters) as $M_2$, and BERT-base (110M parameters) as $M_3$ in this setting.
\textbf{Note that BERT-medium is referred to as $M_2$ in this setting as it is the second model in cascading setup unlike the $K=2$ setting (\ref{subsection_k_2}) in which it was $M_1$}.

\subsubsection{Results and Analysis} 
Figure \ref{fig:acc_cost_curves_K_3} shows the accuracy-cost curves of two cascading approaches: MaxProb (in blue) and Random Baseline (in black) and Table \ref{tab:auc_K_3} compares AUC values achieved by various cascading approaches.
In general, cascading achieves larger improvement (in magnitude) in K=3 setting than K=2 setting.

\paragraph{Efficiency Improvement: }
The accuracy-cost curves show that the cascading system matches the accuracy of larger models $M_2$ and $M_3$ at considerably lesser respective computation costs.
For example, in case of QQP, cascading system matches the accuracy of model $M_3$ by using just $11.07\%$ of $M_3$'s computation cost and of model $M_2$ by using just $23.53\%$ of $M_2$'s computation cost.
The magnitude of efficiency improvement in this setting is higher than that in the K=2 setting.

\paragraph{Accuracy Improvement: }
Cascading also achieves improvement in the overall accuracy.
For example, on the CB dataset, cascading system achieves $83.93\%$ accuracy that is even higher than the largest model $M_3$.
Similar improvements are observed for other datasets also.

\paragraph{Comparing Cascading Approaches: }
Table \ref{tab:auc_K_3} compares AUC values achieved by various cascading approaches.
DTU clearly outperforms all other cascading methods as it achieves the highest AUC values.
We attribute this to DTU's characteristic of utilizing the entire shape of the output probability distribution and not just the highest probability in computing its confidence.

\paragraph{Contribution of $M_1$, $M_2$, and $M_3$ in Cascade: }
\begin{figure}[t!]
    \centering
    \includegraphics[width=6.1cm]{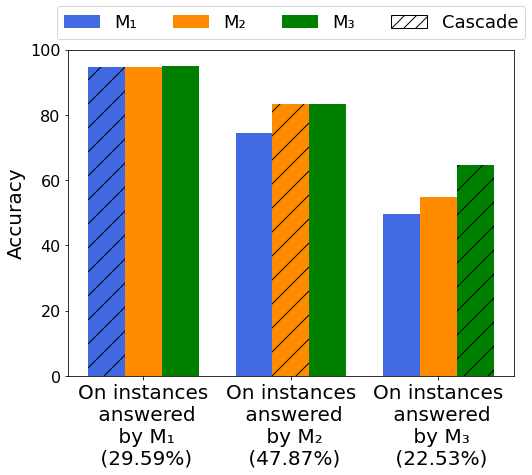}
    \caption{Comparing accuracy of individual models $M_1$, $M_2$, and $M_3$ on the instances answered by each model when used in the cascade for MNLI dataset.}
    \label{fig:contribution_K_3_MNLI}
\end{figure}

Figure \ref{fig:contribution_K_3_MNLI} shows the contribution of individual models $M_1$, $M_2$, and $M_3$ in the cascade for MNLI dataset when the cost is $4.8 \times 10^9$ FLOPs i.e. the point at which accuracy of cascade is equal to that of the largest model $M_3$ (where the horizontal red dashed line drawn from $M_3$ intersects the accuracy-cost curve in Fig \ref{fig:acc_cost_curves_K_3}).
The figure shows that the accuracy of $M_1$ on the instances that were passed to $M_2$ drops by 20.04\% and accuracy of $M_2$ on instances that were passed to $M_3$ drops by 28.53\%.
This shows that the cascading system is good at identifying potentially incorrect predictions of $M_1$ and passes those instances to $M_2$ and similarly good at identifying potentially incorrect predictions of $M_2$ and passes those instances to $M_3$ .

\paragraph{Advantage of introducing another model in the Cascade:}
Comparing figure \ref{fig:contribution_K_3_MNLI} for the K=3 setting with the figure \ref{fig:contribution_K_2_MNLI} for K=2 setting, we find that by introducing a smaller model in the collection, the cascading system can be made more efficient. 
This is because the BERT-medium model answered 78\% instances in K=2 setting and that portion got split into BERT-mini (smaller cost than medium) and medium models in K=3 setting while maintaining the accuracy.
This suggests that the cascading technique utilizes the available models efficiently without sacrificing the accuracy.
% \Neeraj{Compare this with K=2 setting --> M1 of K=2 gives some instances to a smaller model to save cost}
% achieves 10.12\% higher accuracy on those instances. Thus, the system utilizes the models efficiently by using the smaller model $M_1$ for easy instances and the larger model $M_2$ for difficult ones.
We analyze these results for other datasets in Appendix \ref{supp_contribution_mini_medium_base}.

\section{Conclusion and Discussion}
We systematically explored model cascading and proposed several methods for it.
Through comprehensive experiments with $10$ diverse NLU datasets, we demonstrated that cascading improves both the computational efficiency and the prediction accuracy.
We also studied the impact of introducing another model in the collection and showed that it further improves the computational efficiency of the cascading system.
% We also studied the efficacy of cascading in out-of-domain settings.

\paragraph{Selecting Optimal Operating Threshold: }
The selection of confidence threshold for models in the cascade is dependent on the computation budget of the system. 
A low-budget system can select low threshold for the low-cost models (so that low-cost models answer majority of the questions leading to less computation cost) and similarly, high-budget systems can afford to select high thresholds to achieve higher accuracy.
In order to select thresholds in an application-independent manner, the ML's standard practice of using the validation data to tune the hyperparameters can be used. 

\paragraph{Outlier/OOD Detection Techniques: }
Outlier/OOD detection techniques such as \cite{NEURIPS2018_abdeb6f5,Hsu2020GeneralizedOD,NEURIPS2020_f5496252} can also be explored to decide which instance to pass to the bigger models in the cascade.

\paragraph{Including Linear Models in the Cascade:}
This idea can be extended to include non-transformer based less expensive models like linear models or LSTM based models. 
Since the computation cost of these models is significantly lower than the transformer based models and yet they achieve non-trivial predictive performance, a cascading system with these models could achieve even more improvement in computational efficiency. 
We plan to explore this aspect in the future work.

\section*{Limitations}
A potential downside of cascading is that it requires multiple models to be stored. However, we note that the additional space required for models (mini and medium) in K=3 setting is merely $0.44$ times that required for base model (Table \ref{tab:stats}). Thus, it does not pose a serious concern.

\section*{Acknowledgement}
We thank the anonymous reviewers for their insightful feedback. We also thank Tejas Gokhale for giving valuable suggestions. This research was supported by DARPA SAIL-ON and DARPA CHESS programs.

\bibliography{custom}
\bibliographystyle{acl_natbib}

\appendix

\section*{Appendix}

\section{Efficiency in NLP}
With the introduction of large-scale pre-trained language models, the efficiency topic has attracted a lot of research attention. 
Efficiency is being studied from diverse lenses such as training data efficiency \cite{lewis-etal-2019-unsupervised,schick-schutze-2021-exploiting, varshney-etal-2022-unsupervised, Wang2021TowardsZL, mishra-sachdeva-2020-need,ben-zaken-etal-2022-bitfit},
evaluation efficiency 
\cite{rodriguez-etal-2021-evaluation, varshney-etal-2022-ildae}, parameter efficiency tuning methods \cite{li-liang-2021-prefix,pmlr-v97-houlsby19a}, and inference efficiency.
In this work, we focus on inference efficiency and propose model cascading, a simple technique that utilizes a collection of models of varying capacities to accurately yet efficiently output predictions.

\section{Dataset Statistics}
Table \ref{tab:datasets} shows the statistics of all evaluation datasets considered in this work. We consider a diverse set of NLU datasets spanning over several tasks, such as natural language inference, duplicate detection, and sentiment classification.

\section{Cascading with Two Models (K=2)}
\label{supp_sec_k_2}

\subsection{Other Model Combinations}
\subsubsection{Medium and Large}
\begin{figure*}[t]
\centering

    \begin{subfigure}{.3\linewidth}
        \includegraphics[width=\linewidth]{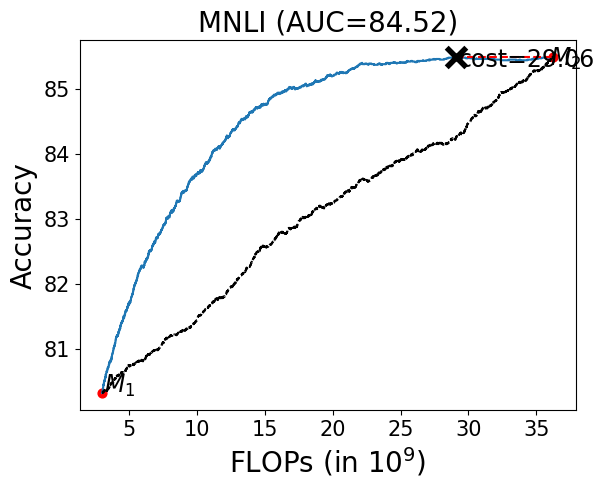}
    \end{subfigure}
    \begin{subfigure}{.3\linewidth}
         \includegraphics[width=\linewidth]{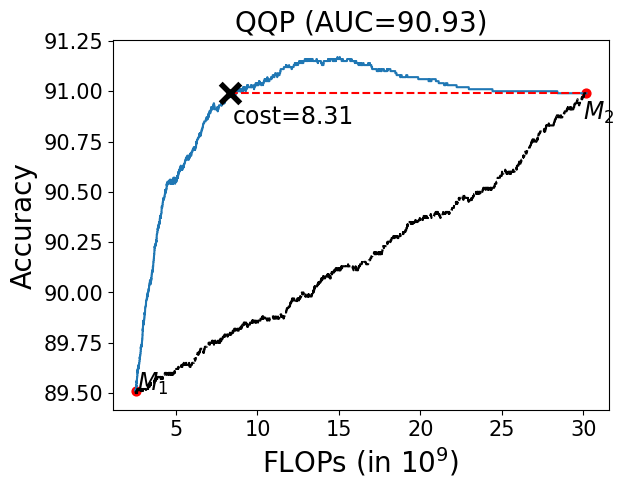}
    \end{subfigure}
    \begin{subfigure}{.3\linewidth}
         \includegraphics[width=\linewidth]{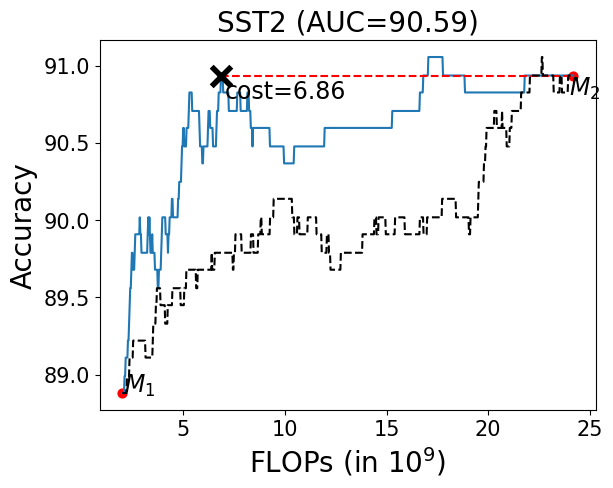}
    \end{subfigure}
    \begin{subfigure}{.3\linewidth}
         \includegraphics[width=\linewidth]{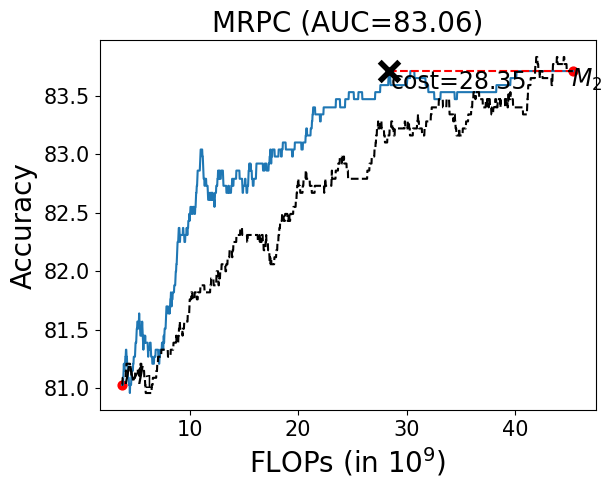}
    \end{subfigure}
    \begin{subfigure}{.3\linewidth}
         \includegraphics[width=\linewidth]{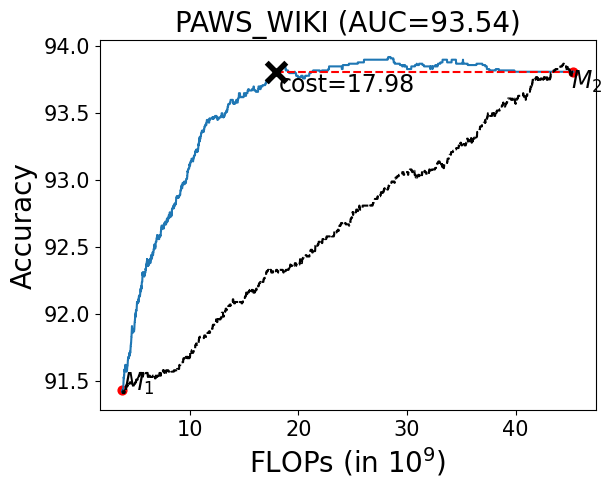}
    \end{subfigure}
    \begin{subfigure}{.3\linewidth}
         \includegraphics[width=\linewidth]{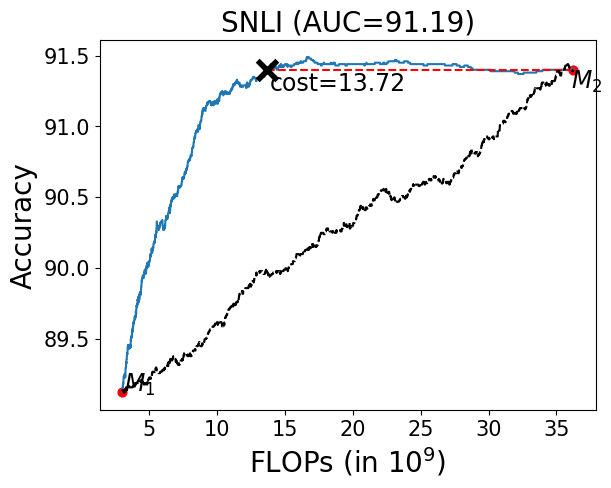}
    \end{subfigure}
    \caption{Accuracy-Cost curves for K=2 setting with $M_1$ as BERT-medium and $M_2$ as BERT-large models.}
    \label{fig:acc_cost_curves_medium_large_K_2}    
\end{figure*}

Figure \ref{fig:acc_cost_curves_medium_large_K_2} shows accuracy-cost curves with MaxProb (in blue) and Random (in black) as cascading approaches with $M_1$ as BERT-medium and $M_2$ as BERT-large.
MaxProb approach clearly outperforms Random approach and achieves considerably higher AUC value.

\subsubsection{Mini and Large}
\begin{figure*}[t]
\centering

    \begin{subfigure}{.3\linewidth}
        \includegraphics[width=\linewidth]{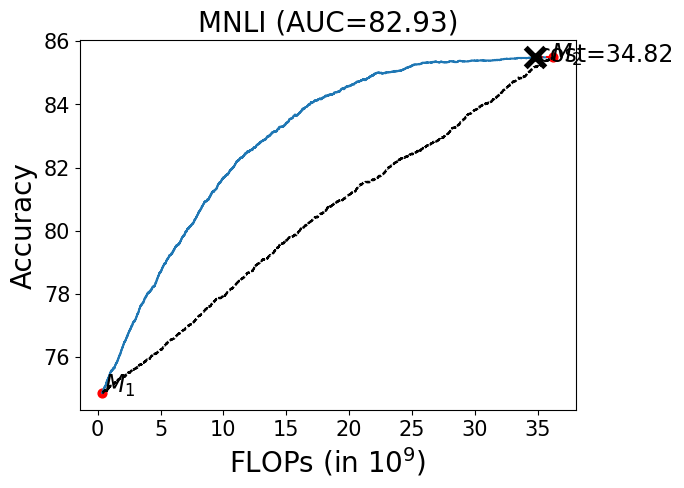}
    \end{subfigure}
    \begin{subfigure}{.3\linewidth}
         \includegraphics[width=\linewidth]{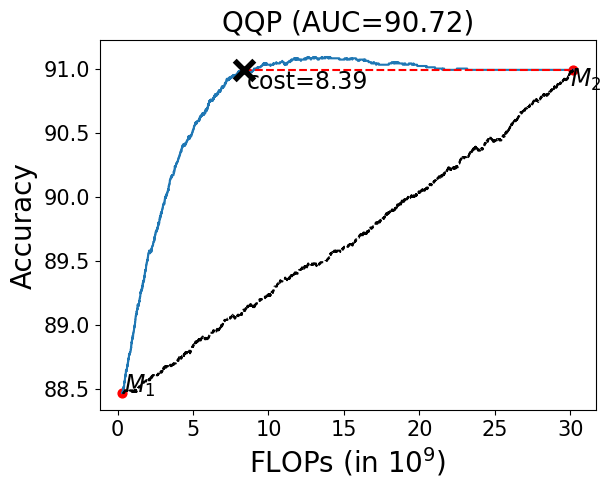}
    \end{subfigure}
    \begin{subfigure}{.3\linewidth}
         \includegraphics[width=\linewidth]{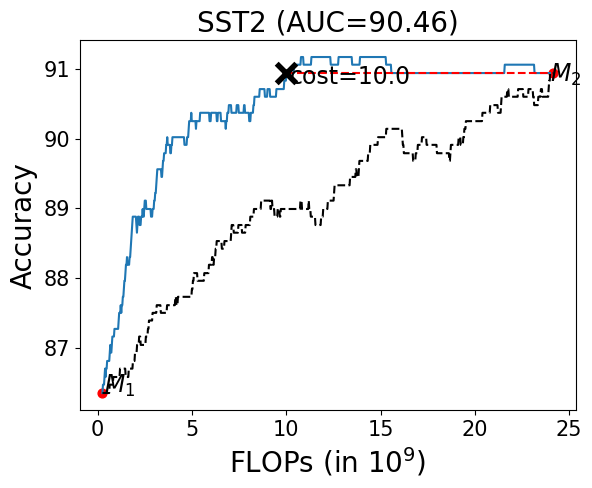}
    \end{subfigure}
    \begin{subfigure}{.3\linewidth}
         \includegraphics[width=\linewidth]{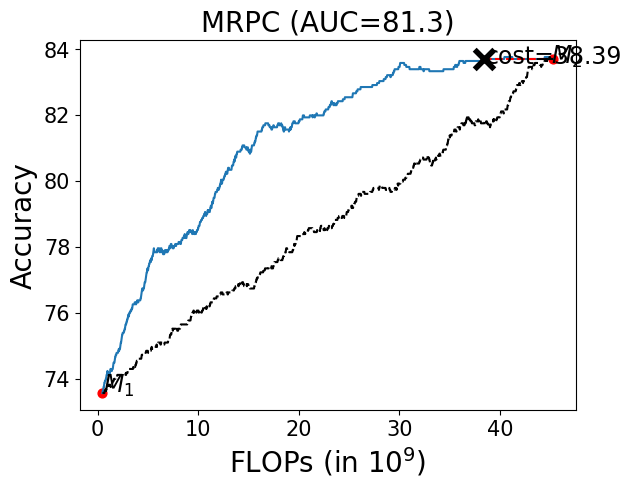}
    \end{subfigure}
    \begin{subfigure}{.3\linewidth}
         \includegraphics[width=\linewidth]{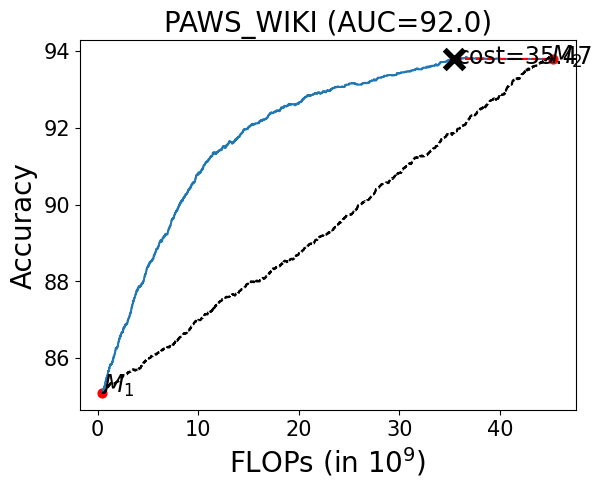}
    \end{subfigure}
    \begin{subfigure}{.3\linewidth}
         \includegraphics[width=\linewidth]{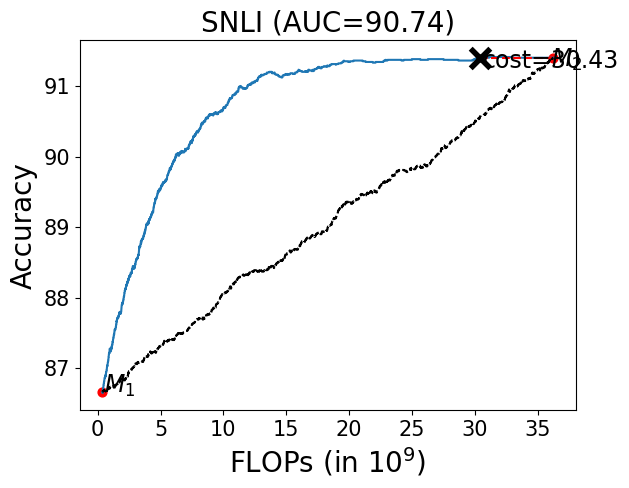}
    \end{subfigure}
    \caption{Accuracy-Cost curves for K=2 setting with $M_1$ as BERT-mini and $M_2$ as BERT-large models.}
    \label{fig:acc_cost_curves_mini_large_K_2}    
\end{figure*}

Figure \ref{fig:acc_cost_curves_mini_large_K_2} shows accuracy-cost curves with MaxProb (in blue) and Random (in black) as cascading approaches with $M_1$ as BERT-mini and $M_2$ as BERT-large.
MaxProb approach clearly outperforms Random approach and achieves considerably higher AUC value.

\subsection{Contribution of $M_1$ and $M_2$ in the Cascade}

\subsubsection{Medium and Base}
\label{supp_contribution_medium_base}
\begin{figure*}[t]
\centering

    \begin{subfigure}{.3\linewidth}
        \includegraphics[width=\linewidth]{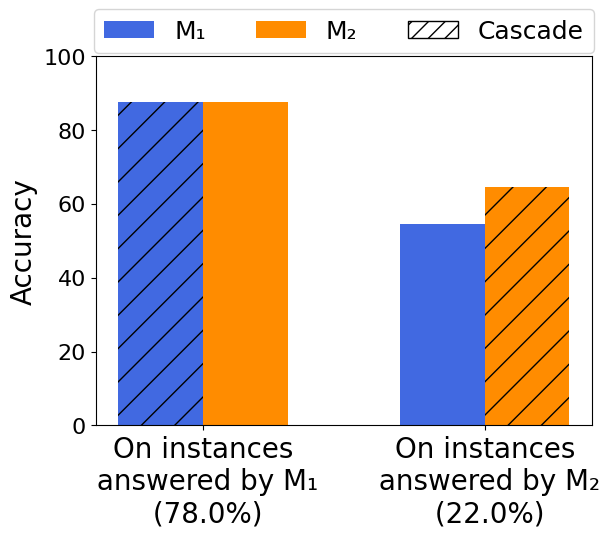}
    \end{subfigure}
    \begin{subfigure}{.3\linewidth}
         \includegraphics[width=\linewidth]{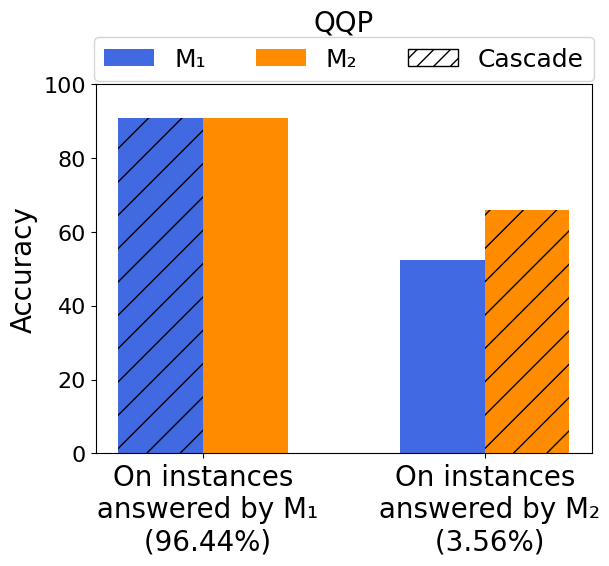}
    \end{subfigure}
    \begin{subfigure}{.3\linewidth}
         \includegraphics[width=\linewidth]{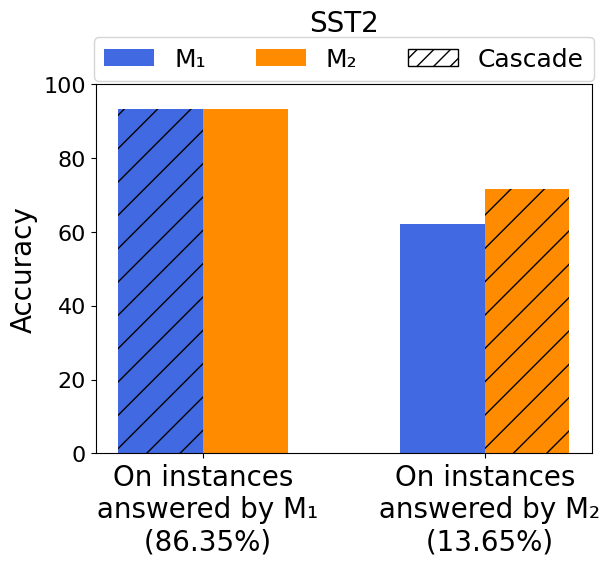}
    \end{subfigure}
    \begin{subfigure}{.3\linewidth}
         \includegraphics[width=\linewidth]{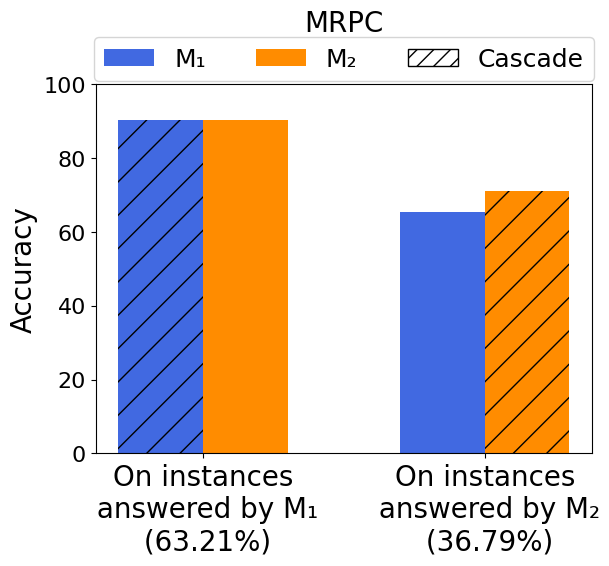}
    \end{subfigure}
    \begin{subfigure}{.3\linewidth}
         \includegraphics[width=\linewidth]{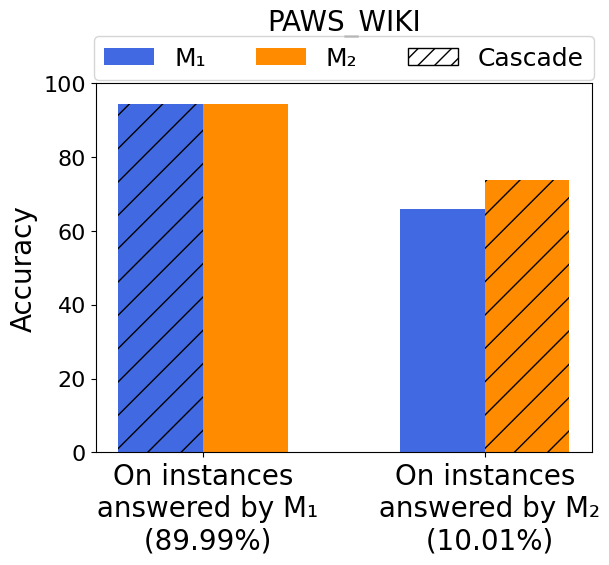}
    \end{subfigure}
    \begin{subfigure}{.3\linewidth}
         \includegraphics[width=\linewidth]{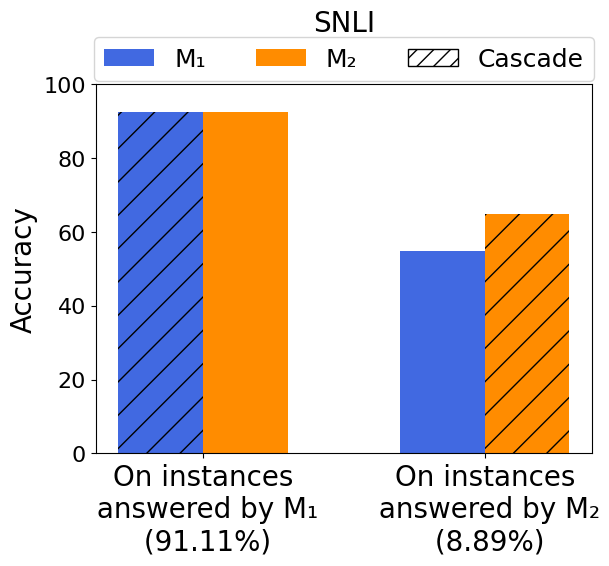}
    \end{subfigure}
    \caption{Contribution of $M_1$ and $M_2$ for K=2 setting with $M_1$ as BERT-medium and $M_2$ as BERT-base.}
    \label{fig:contributions_K_2}    
\end{figure*}

Figure \ref{fig:contributions_K_2} shows the contribution of individual models $M_1$ and $M_2$ in the cascade when accuracy of $M_2$ is same as that of cascading system.
We analyze this for MNLI dataset in section \ref{subsection_k_2} and provide figures for a few other datasets here.

\section{Cascading with Three Models (K=3)}

\subsection{Contribution of $M_1$, $M_2$, and $M_3$ in the Cascade}

\subsubsection{Mini, Medium, and Base}
\label{supp_contribution_mini_medium_base}
\begin{figure*}[t]
\centering

    \begin{subfigure}{.3\linewidth}
        \includegraphics[width=\linewidth]{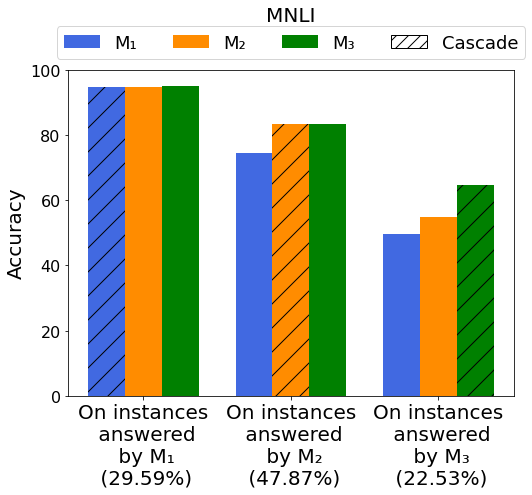}
    \end{subfigure}
    \begin{subfigure}{.3\linewidth}
         \includegraphics[width=\linewidth]{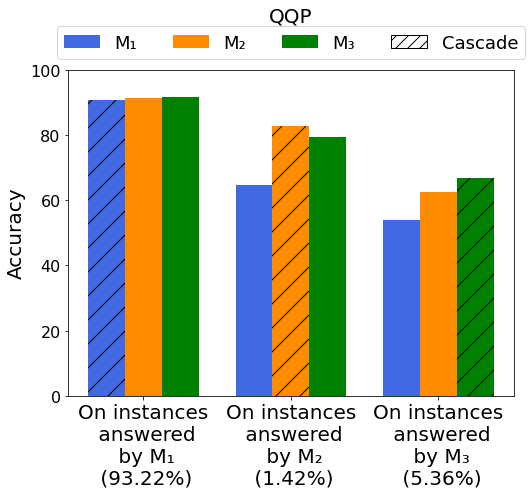}
    \end{subfigure}
    \begin{subfigure}{.3\linewidth}
         \includegraphics[width=\linewidth]{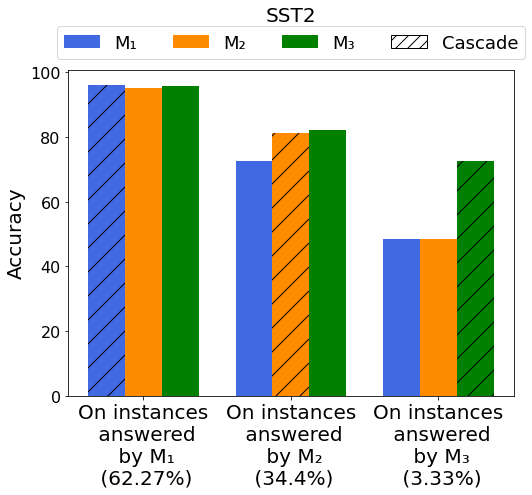}
    \end{subfigure}
    \begin{subfigure}{.3\linewidth}
         \includegraphics[width=\linewidth]{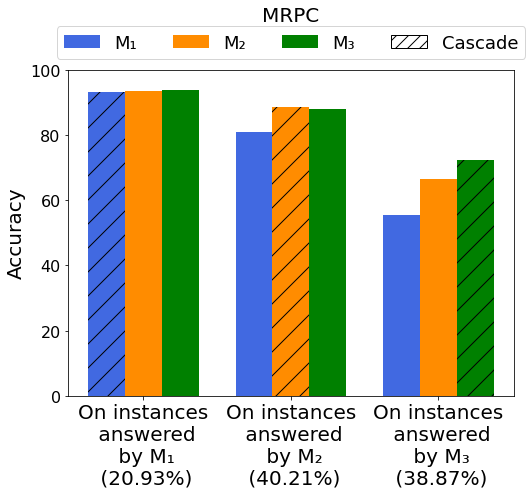}
    \end{subfigure}
    \begin{subfigure}{.3\linewidth}
         \includegraphics[width=\linewidth]{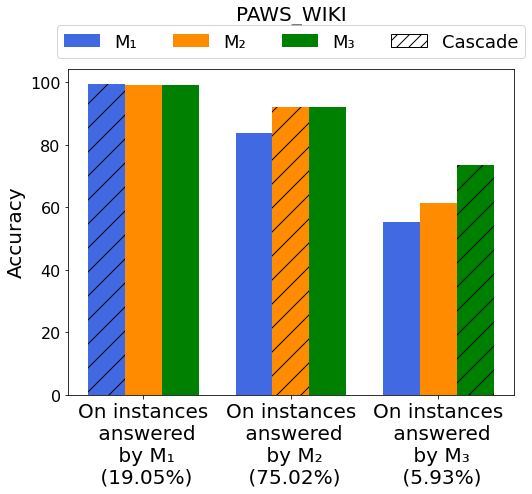}
    \end{subfigure}
    \begin{subfigure}{.3\linewidth}
         \includegraphics[width=\linewidth]{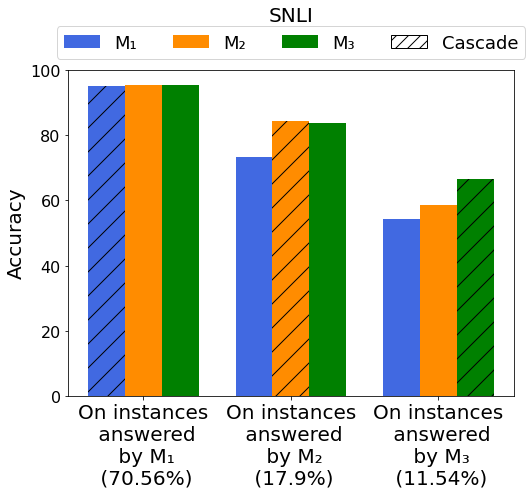}
    \end{subfigure}
    \caption{Contribution of $M_1$, $M_2$, and $M_3$ for K=3 setting with $M_1$ as BERT-mini, $M_2$ as BERT-medium, and $M_3$ as BERT-base.}
    \label{fig:contributions_K_3}    
\end{figure*}

Figure \ref{fig:contributions_K_3} shows the contribution of individual models $M_1$, $M_2$, and $M_3$ in the cascade when accuracy of $M_3$ is same as that of cascading system.
We analyze this for MNLI dataset in section \ref{subsec_K_3} and provide figures for a few other datasets here.

\begin{table}
\small
    \centering
    \begin{tabular}{p{1.7cm}p{1cm}|p{1.8cm}p{0.9cm}}
     \toprule
        \textbf{Dataset} & \textbf{Size} & \textbf{Dataset} & \textbf{Size}\\
         \midrule
        MNLI & 19645 & QNLI & 5650 \\
        QQP & 40371 & SST2 & 872 \\ 
        COLA & 1042 & CB & 56 \\ 
        DNLI & 16396 & MRPC & 1639 \\ 
        PAWS & 7994 & SNLI & 9840 \\ 
        
        \bottomrule
    \end{tabular}
    \caption{Statistics of evaluation datasets considered in this work.}
    \label{tab:datasets}
\end{table}

\subsection{Overall Efficiency and Accuracy Improvement}
\begin{figure*}[t!]
    \centering
    \begin{subfigure}{0.48\linewidth}
        \includegraphics[width=\linewidth]{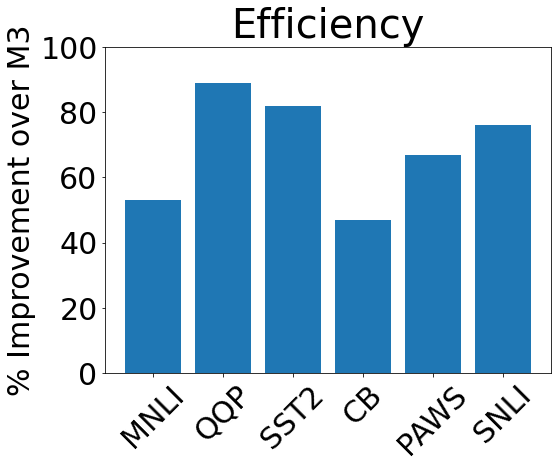}
    \end{subfigure}
    \begin{subfigure}{0.48\linewidth}
         \includegraphics[width=\linewidth]{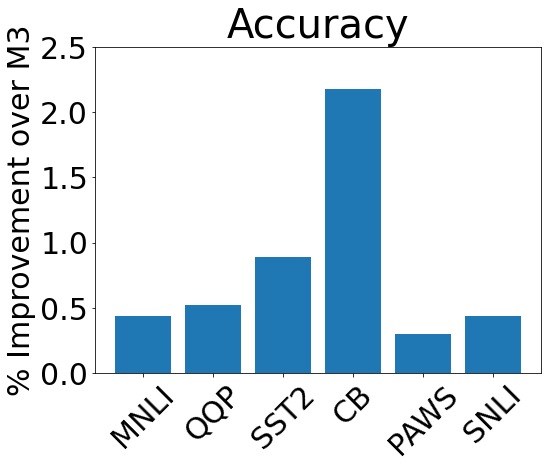}
    \end{subfigure}
    
    \caption{Efficiency and accuracy improvement achieved by the cascading system (using \textit{DTU} method (\ref{subsec_approaches})) over the largest model $M_3$ in K=3 setting.}
    \label{fig: overall_improvement_K_3}    
\end{figure*}

Figure \ref{fig: overall_improvement_K_3} (left) illustrates efficiency improvements achieved by a cascading method over the largest model ($M_3$) in K=3 setting.
For example, in case of QQP dataset, the cascading system achieves $88.93\%$ computation improvement over $M_3$ i.e. it requires just $11.07\%$ of the computation cost of model $M_3$ to attain equal accuracy.
Then, we show that cascading also achieves improvement in prediction performance.
Figure \ref{fig: overall_improvement_K_3} (right) illustrates the accuracy improvements achieved over $M_3$ in K=3 setting.
For example, on CB dataset, the cascading system achieves $2.18\%$ accuracy improvement over $M_3$.

\end{document}